\RequirePackage{fix-cm}

\documentclass[smallextended]{svjour3}       % onecolumn (second format)
\usepackage{epsfig}
\usepackage{calc}
\usepackage{amssymb}
\usepackage{amstext}
\usepackage[font=small,labelfont=bf,tableposition=top]{caption}% per avere le didascalie diverse tra tabular e figura in una table
\usepackage{multicol}
\usepackage{pslatex}
\usepackage{apalike}
\usepackage{mathtools, cuted}
\usepackage{url}
\usepackage{listings}
\usepackage{caption}
\usepackage{color}
\usepackage{algorithm, algpseudocode} % pseudocodice
\usepackage{natbib}
\usepackage[skip=1ex]{caption}

\DeclareMathOperator*{\argmax}{arg\,max}

\begin{document}

\title{
A fast algorithm for complex discord searches in time series: HOT SAX Time
}

\author{Paolo Avogadro        \and
        Matteo Alessandro Dominoni %etc.
}

% \institut{Paolo Avogadro \at    \email{paolo.avogadro@unimib.it}  
%            \and
%            Matteo Alessandro Dominoni \at  \email{matteo.dominoni@unimib.it}
%            }
% %           
 \institute{%
            Paolo Avogadro \at    \email{paolo.avogadro@unimib.it} %\phone{+39-02-6448-7891}  
            \and
            Matteo Alessandro Dominoni \at  \email{matteo.dominoni@unimib.it} \\
              Universit{\`a} degli studi di Milano-Bicocca, \\ 
              viale Sarca 336, 22126, Milano, Italy, \\
              Tel.: +39-02-6448-7891                 
 }

\maketitle

\begin{abstract}
Time series analysis is quickly proceeding towards long and complex tasks.
In recent years, fast approximate algorithms for discord search have been proposed in order to compensate for the increasing size of the time series. It is more interesting, however, to find quick exact solutions. 
In this research, we improved HOT SAX by exploiting two main ideas: the warm-up process, and the similarity between sequences close in time. The resulting algorithm, called HOT SAX Time (HST), has been validated with real and synthetic time series, and successfully compared with HOT SAX, RRA, SCAMP, and DADD. 
The complexity of a discord search has been evaluated with  a new indicator, the cost per sequence (\textit{cps}), which allows one to compare searches on time series of different lengths. Numerical evidence suggests that two conditions are involved in determining the complexity of a discord search in a non-trivial way: the length of the discords, and the noise/signal ratio. 
In the case of complex searches, HST can be more than 100 times faster than HOT SAX, thus being at the forefront of the exact discord search. 
\end{abstract}


%\onecolumn \maketitle \normalsize 
\setcounter{footnote}{0} \vfill

%\section{\uppercase{Introduction}}
%\label{sec:introduction}
%\section{\uppercase{Manuscript Preparation}}
%\section*{\uppercase{Acknowledgements}}

%\section{Introduction}\label{sec:intro}

\section{Introduction and related Works}\label{sec:related}

Anomaly discovery in time series is an active research field \citep{chandola} \citep{gupta2014} where many approaches are taken into consideration. 

One of the first steps when analysing a time series is to reduce its dimensionality. Symbolic aggregate approximation (SAX) \citep{sax} is a successful algorithm which goes in this direction associating each sequence to a symbolic sequence. Later on, an improved version, called iSAX, was presented allowing for better clusterizations \citep{iSAX}.

Discords were introduced in 2005 \citep{hotsax}, where it was also suggested a fast algorithm for finding them: HOT SAX.
The use of Haar wavelets \citep{haar2} improved the pruning power in respect to SAX.
HOTiSAX \citep{hotIsax} is an algorithm based on iSAX which is up to four times faster than HOT SAX.
The most advanced methods for a complete characterization of the time series are those of the Matrix Profile series initiated in 2016 \citep{matrixI}. The algorithms of the Matrix Profile provide a quick calculation of the distance among all the sequences and as a result they can also return discords, however given this broad nature their complexity grows quadratically with the length of the time series.
The research state of the art has been progressing in order to take into account for the increasing length of the time series and complexity of the tasks. We are now in the hundred million points era, for example regarding motif analysis \citep{matrixII} and \citep{gao}.

If the size of the time series is such that the RAM memory is not enough, one needs to keep at least part of the data on disk, and for this purpose it was developed the Disk-Aware Discord Discovery (DADD, also known as DRAG) algorithm \citep{DADD}. %, which has a complexity O($N$).
MERLIN is a new algorithm based on DADD which can quickly scan all the  discords within a given length range \citep{MERLIN}.

A first route that can be followed in order to speed up the calculation regards parallelizing the existing algorithms, as was done for example by \cite{russi2019}, where a version of HOT SAX has been developed for Intel Many-Core Systems. The fastest algorithm involving graphic card accelerators is SCAMP of the Matrix Profile series \citep{scamp}.  
A different direction involves developing approximate discord algorithms.
Rare Rule Anomaly \citep{grammarviz2}, for example, exploits the Kolmogorov complexity of the SAX symbolic sequences to extract anomalies close to the definition of discords.
At variance, a recent approximate discord search method \citep{segment}, based on segmentation and clustering, automatically suggests the size of the anomaly, and it is faster than HOT SAX. The overlap between these anomalies and discords is good, however if one searches for the first $k-$anomalies there is an increasing probability to get wrong results.
All the algorithms which return approximate discords have this common drawback: the price to pay for an improved speed is a drop in accuracy in respect to an exact discord search. 

Although there are many articles regarding discord search, there has not been an important attempt to understand why the heuristics perform well on some time series, while the results are poorer on others. 
It is thus desirable to obtain a general view of the problem for better understanding the nature of the anomalies. As a result, one could obtain new ideas regarding time series and anomaly detection. 
Our methodology was to understand in detail the mechanism at the basis of HOT SAX and check if the neighborhood properties of the times series might help in finding shortcuts. Once we found areas where HOT SAX could be improved, we implemented the new functions and we checked if these ideas were able to produce significant improvements. We used HOT SAX as a comparison for our tests since it is a benchmark on the subject; we also included DADD and SCAMP, and a fast freely available approximate algorithm: RRA. 
Notice that our algorithm has not yet been parallelized, so all the comparisons are limited to the state of the art serial codes.
%
%%%%%%%%%%%%%%%%%%%%     NEW     %%%%%%%%%%%%%%%%%%%
During the evaluation process, the great diversity in computational costs between different discord searches became particularly striking.
In order to disambiguate the roles of the different parameters of a discord search, 
%%%%%%%%%%%%%%%%%%%%  FINE NEW   %%%%%%%%%%%%%%%%%%%
we defined a quantity: the ``cost per sequence'' (\textit{cps}) (detailed in Sec. \ref{ssec:cps}) 
%%%%%%%%%%% NEW %%%%%%%%%%%%
The \textit{cps} allows one to compare discord searches on time series of different length and rate them in terms of complexity.
%%%%%%%%%% end NEW %%%%%%%%%%%%%%%%%%%%%%%%%%%%%
%%%%%%%%%%%%%%%%%%%% NEW %%%%%%%%%%%%%%%%%%%%%%%%%%%
Following this new definition, we were able to notice that HST is particularly faster than HOT SAX at finding discords for complex searches.
%%%%%%%%%%%%%%%%%%%% END NEW %%%%%%%%%%%%%%%%%%%%%%%
\\
The rest of the paper is organized as follows:
\begin{itemize}
\item After clarifying the notations and defining useful quantities, Sec. \ref{sec:background} provides a brief explanation of the brute force approach and of HOT SAX.
\item Sec. \ref{sec:vhs} details HST and the rationale of the modifications in respect to HOT SAX.
\item Sec. \ref{sec:validation} compares the execution speeds of HST and its main available competitors. We introduce a new indicator: the cost per sequence of a discord search and we detail cases where HST is particularly fit for the job. 
\item Sec. \ref{sec:conclusions} contains the conclusions and the future work.
\end{itemize}

\section{Theoretical background} \label{sec:background}

\subsection{Terminology}\label{ssec:terminology}
\begin{itemize}
 
 \item The points of a time series are denoted with $p_j$, where the subscript indicates the time.

  \item A sequence, containing $s$ points, is described with the time of its	 first point. For example, the points of sequence $k$ are:  $p_k, p_{k+1}, ..., p_{k+s-1}$. In this paper, unless specifically stated, we consider z-normalized sequences \citep{Keogh2003}.
 
  \item If the number of points of a time series is denoted with $N_{tot}$, the number of sequences of length $s$ therein is smaller, and denoted with the letter $N=N_{tot}-s+1$. The last sequences are not complete and so they are removed from the search space.
 
  \item The Euclidean distance among two sequences $k$ and $l$ is:
 \begin{equation}\label{eq:euclideanDistance}
   d(k, l)  = \sqrt{ \sum_{i=0}^{s-1} \left(  p_{k+i} - p_{l+i} \right)^2  }.
 \end{equation}

 The process of z-normalization implies a large memory expenditure.
It is possible to save memory by storing the averages and standard deviations of all of the sequences and proceed with the following function instead:
  \begin{equation}\label{eq:zDist}
   d(k, l)  = \sqrt{ \sum_{i=0}^{s-1} \left(  \frac{ p_{k+i} -\mu_k} {\sigma_k} - \frac{ p_{l+i} -\mu_l}{\sigma_l}  \right)^2  },
 \end{equation}
as pointed out by \cite{matrixII}, the same result can be obtained with the help of the scalar product: 
 \begin{equation}\label{eq:dotDistance}
 d(k,l)    = \sqrt{ 2s \left(  1  - \frac{ k\cdot l  - s \mu_k \mu_l }{ s \sigma_k \sigma_l }   \right),   }  
 \end{equation}
the notation $k\cdot l $ is associated with the scalar product between the two sequences (vectors), $s$ is the length of the sequences,
$\mu_k$ and  $\sigma_k$ are the mean and standard deviation of sequence $k$.

  \item The nearest neighbor distance ($nnd$) for sequence $k$ is obtained as: 
 \begin{equation}
  nnd(k) =  \min_{j:|k-j|\ge s} d(k, j).  
 \end{equation}
The search space for the minimization includes all the sequences, $j$, of the times series, with the exception of those that overlap ($j:|k-j| \ge s$) with $k$.  This non self-match condition serves in order to avoid spurious low values of $nnd$ due to overlaps. 

  \item The quantity $ngh(i)$ returns the position of the nearest neighbor of sequence $i$: 
  $$
  nnd(i)=d\left(i, ngh(i) \right).
  $$
  \item The $nnd$ profile is the set of the current $nnd$s of all the sequences of a time series. In practice many of the $nnd$s returned by HST are only approximate values. Had these values been exact, the $nnd$ profile would be identical to a special case of \textit{Matrix Profile}: the self-similarity join profile, or $P_{AA}$ \citep{matrixI}.

  \item In order to highlight that some quantities are stored in an array or variable of the code, we use typewriter characters. For example \verb|nnd[i]|, or \verb|ngh[i]| are the arrays containing the information regarding the $nnd(i)$ and $ngh(i)$.  
  \item In the rest of the paper we will refer to time-distance between two sequences $i$, $k$ as $|i-k|$. For example the nearest time-neighbors of sequence $i$ are $i-1$ and $i+1$.
  \item In order to compare the performance of two algorithms we can resort at using the distance-speedup (D-speedup) calculated as the ratio of the number of distance calls between the algorithms. We can also use the time-speedup (T-speedup) obtained as the ratio of the runtimes of the two algorithms on the same dataset for the same task.
  
\end{itemize}

\subsection{Discord}\label{ssec:HS}
The concept of discord follows an intuitive idea regarding anomaly search in time series. A sequence of length $s$ can be considered as an $s$-dimensional vector (or point of an $s$-dimensional space). With this view, an isolated point is an anomaly.
In practice, the discord is defined as the sequence with the highest $nnd$ in respect to all the other sequences of the time series.
$$
discord = \argmax_i \left( \frac{}{}  nnd(i) \right)
$$

%%%%%%%%%%%%%%%%%%%%%  NEW %%%%%%%%%%%%%%%%%%%
The second discord is defined as the sequence with the highest $nnd$ as long as it does not overlap the first discord.
The $k$-th discord is the sequence with the highest $nnd$ as long as it does not overlap any of the previous $k-1$ discords.

We will often call a sequence a  \textit{good discord candidate} if it has the highest $nnd$ after its distance has been calculated with all the others during the execution of the algorithm. According to this definition, the last sequence which becomes a \textit{good discord candidate} will be the discord.

%%%%%%%%%%%%%%%%%%%%% END NEW %%%%%%%%%%%%%%%%

%
\subsection{Brute Force approach}\label{ssec:brute}
A brute force algorithm for discord search requires two nested loops:
\begin{enumerate}
 \item The external one runs on all the sequences of the time series. It is a maximization procedure for finding the sequence with the highest $nnd$.
 \item The internal loop provides the minimization procedure for obtaining  the nearest neighbor distance of each sequence. This process involves the calculation of the distances between the selected sequence and all the others (excluding self-matches).
\end{enumerate}
As a result, the complexity of a brute force calculation as a function of the number of sequences $N$ is  $O(N^2)$.
\subsection{HOT SAX}\label{ssec:hotsax}
HOT SAX shows a way to skip a large part of the distance calculations of a brute force approach. Via the symbolic aggregate approximation (SAX) \citep{sax} it is possible to quickly assign every sequence to a much shorter symbolic sequence (or SAX cluster). This dimensionality reduction procedure clusterizes sequences efficiently.  Because of the properties of SAX, sequences belonging to the same SAX cluster can also be Euclidean neighbors, and it is possible to exploit this fact for skipping most of the calculations.		
In order to improve the brute force approach, the outer and inner loops are re-arranged following the indications obtained with SAX. The external loop is now structured according to the SAX clusters: from the smallest to the biggest ones (containing more sequences). The order of the internal loop is dynamic, depending on the current sequence of the external loop. If the external loop has arrived on a sequence belonging to a given cluster, the sequences of that cluster are set at the beginning of the internal loop. 
The remaining part of the inner loop follows a pseudo-random order. 
At any point in the inner loop, as soon as the $nnd$ of the candidate sequence becomes smaller than the best so far value, the rest of the inner loop can be skipped (since the sequence under observation cannot be a discord). 
The rationale of these choices is that small clusters, containing only a few sequences, are good candidates for finding ``isolated sequences'' or sequences for which the $nnd$ is high. 
Moreover, close Euclidean neighbors are likely found in the same cluster of a sequence. If one finds a close neighbor of a sequence, its approximate $nnd$ can become smaller than the current best (highest) one, allowing one to skip the remaining distance calls of the inner loop. Those calls, in fact, can only lower than the actual $nnd$ value of the sequence.

\section{HOT SAX Time}\label{sec:vhs}
\subsection{The model to be improved} \label{ssec:model} 
In the following, we will detail how to obtain the new algorithm from HOT SAX and the origin of these modifications. %

In order to improve HOT SAX we need a model regarding discords and their search.
Let's recall the discord definition as the sequence with the highest $nnd$ value. Following this definition, one can think of a discord search as a way to find the maximum of a profile when the profile is not known beforehand.
An intuitive way to see HOT SAX is the following. SAX provides a sort of hazy vision of the $nnd$ profile where one can see that there are peaks (small clusters) but it is not possible to distinguish their height. At this point one has an idea regarding where to search the discords. 
The inner loop clarifies this view by returning an approximate $nnd$ value for each of the sequences. If the sequence goes through the whole inner loop, its $nnd$ is exact and the sequence is a good discord candidate. However, the cost for each one of these ``perfect clarifications'' (the number of distance calls) is essentially identical to the number of sequences the time series. The total cost of the ``clarification'' process determines how difficult it is to find the discord. If one is sure that the approximate $nnd$s of all the other sequences are lower than the best so far candidate, that sequence is the discord.

In the rest of the paper we will follow the same general mechanism, but we will try to improve the ``view'' which guides the  search process. 
With the help of the similarity between time-close sequences, we will be able to obtain a better idea of the position of the discords with an indication of their $nnd$s. At this point, the search will become easier since most of the sequences, having a low $nnd$, will be discarded immediately.

\subsection{Speed-up  the k-th discord}\label{ssec:speedK}
%%%%%%%%%%%%%%%%%%%%%%% NEW %%%%%%%%%%%%%%%%%%%%%%%%%%%%
%%%%%%%%%%% NEW %%%%%%%%%%%%
Here we detail 
a well-known technique \citep{haar2} that can be used to diminish the number of calculations for the $k$-th discord (once the first $k-1$ discords have been found). We will use it later to find a good $nnd$ profile at a low computational cost.
%%%%%%%%%%%%%%%%%%%%%%% END NEW %%%%%%%%%%%%%%%%%%%%%%%%

It is useful to remind the concept of $k$-th discord, which is the sequence with the highest $nnd$ non-overlapping any of the previous $k-1$ discords.
During the calculation of each discord, the code should update the approximate $nnd$ values for all the sequences. These quantities are obtained by refreshing the $nnd$s in the inner loop.
% of the algorithm. 
The memory cost of keeping track of the approximate $nnd$s is just the size of the time series O($N$).
Although these approximate $nnd$s are rough estimates of the exact ones, they are very useful, since the real anomalies (discords) have $nnd$ values usually quite higher than those of the ``common'' sequences.
During the search for the second discord, even before starting the inner loop for a sequence, one should check its current approximate $nnd$. If the sequence under consideration exhibits an $nnd$ value lower than the present best value, the whole inner loop of that sequence can be skipped (since it is a minimization). Thanks to this simple procedure, it is possible to skip most of the calculations for the $k$-th discord.  
%%%%%%%%%%%%%%%%  NEW %%%%%%%%%%%%%%%%%%%%%%%%%%%%%%%%%%
In the next section, we provide a solution that allows applying the aforementioned technique also in the case of the first discord.
%%%%%%%%%%%%%%%% END NEW %%%%%%%%%%%%%%%%%%%%%%%%%%%%%%%

\subsection{Warm-up}\label{ssec:warmup}
%%%%%%%%%%%%NEW %%%%%%%%%%%
In this section we explain the first main procedure at the basis of HST.
%%%%%%%%%%% END NEW %%%%%%%%%%%
%
\begin{figure}[ht]
 \includegraphics[width=0.49\textwidth]{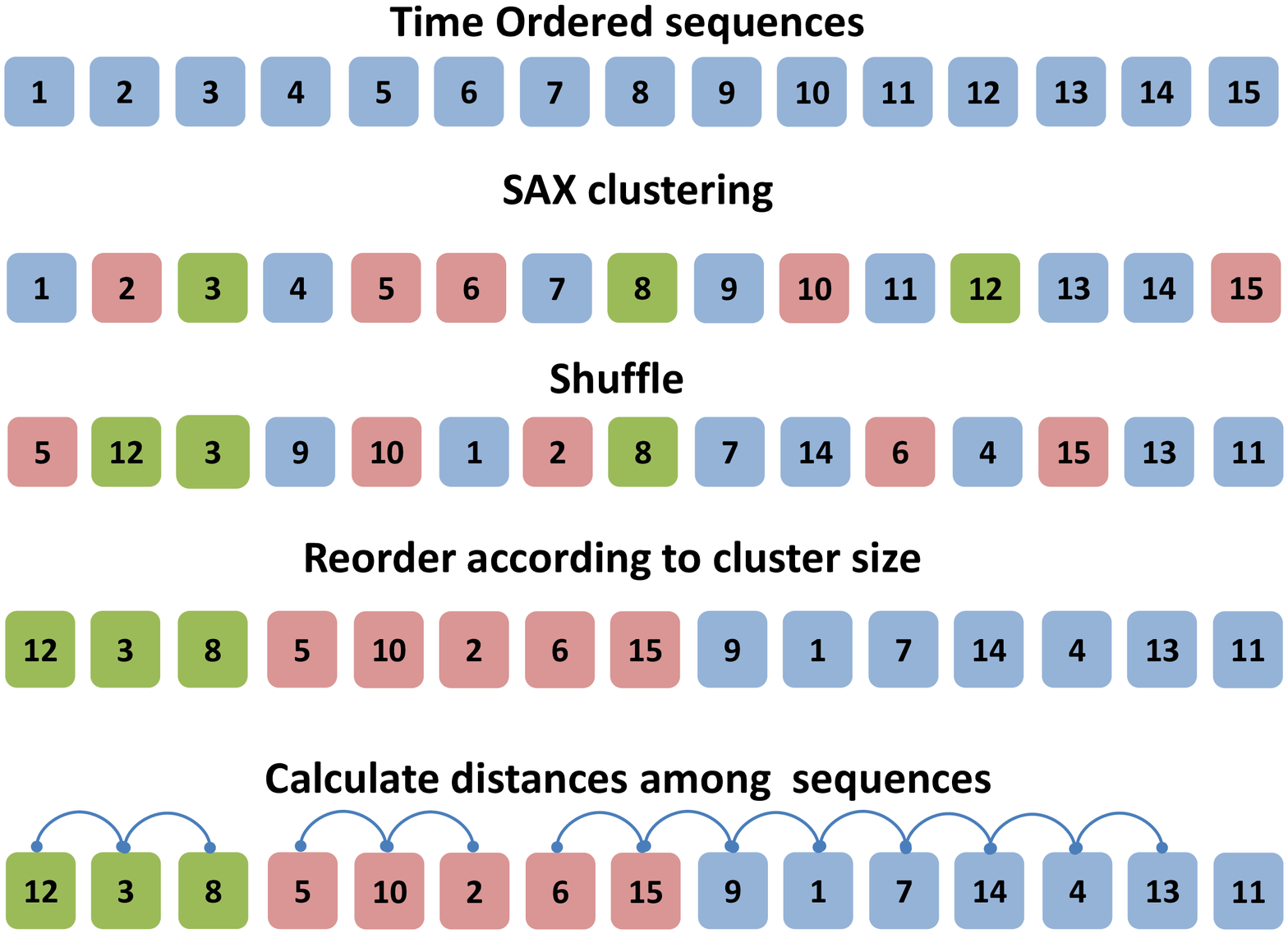}
 \includegraphics[width=0.49\textwidth]{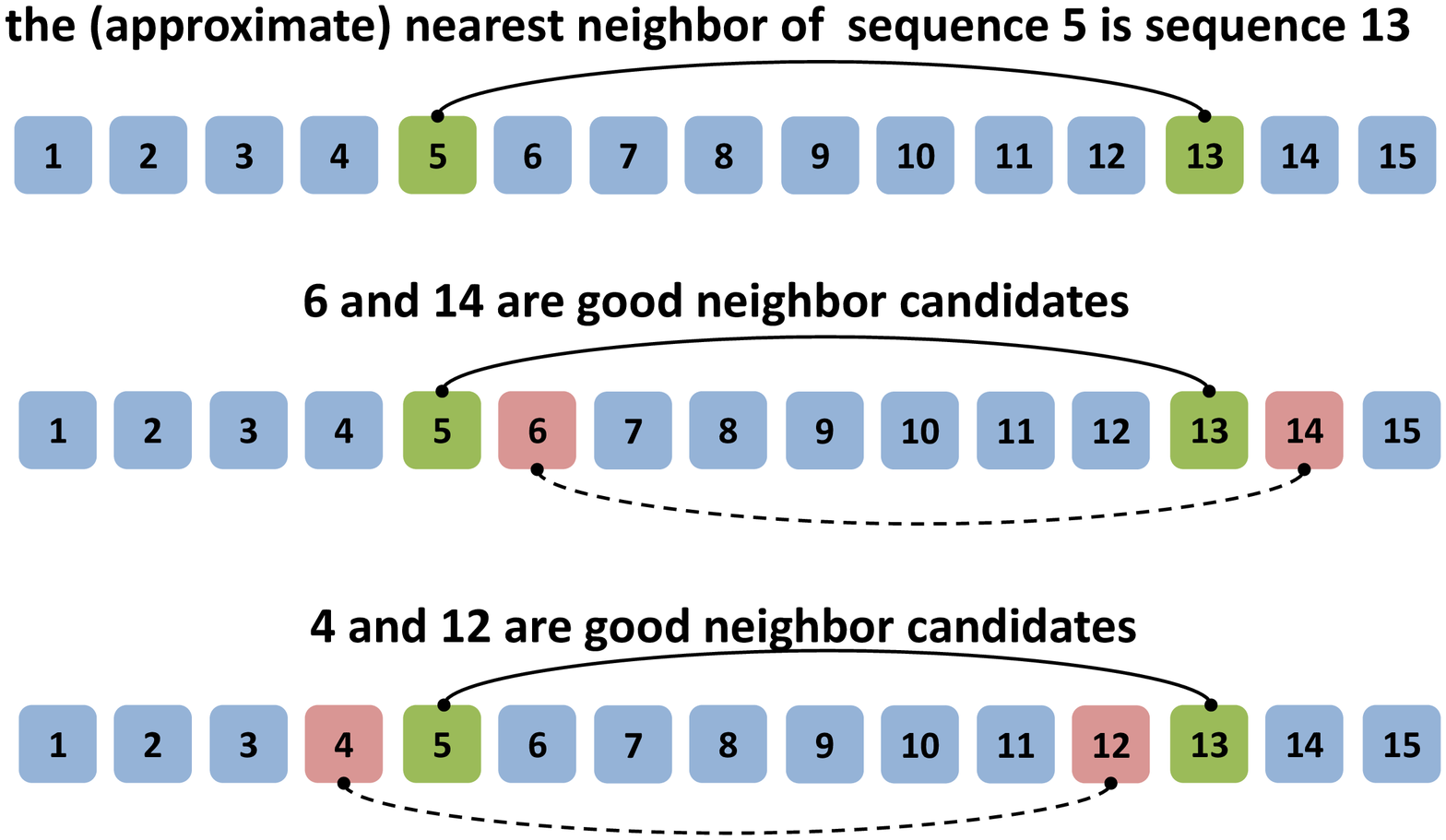}
 \caption{(Left) At the beginning, all the sequences are simply ordered according to their first point. We suppose that the length of the sequences is $5$, we show here only the beginning point of each sequence.  As a first step, the SAX procedure assigns each sequence to a cluster (in the example there are 3 clusters, defined by color: green, blue, and pink).
 The sequences are then shuffled in order to avoid that many close sequences are one after the other. The sequences are then grouped according to the cluster size: at the beginning the green cluster containing 3 sequences, then the pink one and the blue one. The final step consists in calculating the distance between adjacent sequences (according to this new order).  Because of the self-match condition the following distances cannot be calculated: $d(8,5)$, $d(2,6)$, and  $d(13,11)$.
 In the case of adjacent clusters, the distance is calculated between the last sequence of one cluster and the first one the next cluster. For example, the last sequence of the pink cluster, $15$, is connected with the first one of the blue cluster, $9$. Notice that for sequence $11$ it is not possible to calculate any distance, and for this reason, a default $nnd$ high value is assigned.
 (Right). 
 The time topology suggests where to find good neighbor candidates. If we know that sequence 5 and 13 are neighbors, then 6 and 14 are likely good neighbors (also 4 and 12).
 } \label{fig:warmup}
\end{figure}
It would be particularly interesting to exploit the idea of the previous section also at the beginning of the search. 
Unfortunately, at the time of the calculation of the first discord there is no available approximate $nnd$ value for all the sequences. We now detail a fast procedure to build such an approximate $nnd$ profile. 
The idea is to find a new order where similar, non overlapping sequences are one after another.
Once this new ordering has been obtained one can perform a chain of calls to the distance function from one sequence to the next one and so on. As a result, for each sequence there are two distance calls to two ``neighbors'' and thus it is possible to obtain a rough $nnd$ profile. The total number of distance calls for this procedure is essentially equal to the number sequences of the time series. 

Since the rationale of HOT SAX is that sequences belonging to the same cluster are also likely Euclidean neighbors, it becomes natural to calculate the distance between two sequences within the same cluster. 

Let's consider a situation in which all the sequences have already been grouped in clusters by SAX.
There can be different variations regarding how to retrieve the sequences from the SAX clusters, and for some of them, the temporal order of the sequences is preserved. This can be a problem since we want to obtain a chain of distance calls, and time consecutive sequences are self-matches (for which the distance should not be calculated). In order to avoid this problem, it is useful to shuffle randomly the sequences. With this simple procedure one can avoid many of the possible self-matches and increase the number of ``valid'' distance calls. Sometimes the randomization procedure cannot avoid self-matches. For example, if a cluster contains only 3 sequences (of length 10) beginning at time 67, 73, and 75,  non-self-match distance calls among them  are impossible (independent of their order). Before the warm-up procedure we initialize all of the $nnd$s with a very high value. With this choice, no possible discord candidate is neglected. 
%
%%%%%%%%%% NEW %%%%%%%%%%%%
The last sequence of a cluster is coupled with the first one of the next cluster.
%%%%%%%%%% END NEW %%%%%%%%

To summarize, the warm-up procedure  (see Fig. \ref{fig:warmup}, left) consists of 3 steps (and 2 pre-requisites: the initialization of the $nnd$s, and the SAX procedure):  
\begin{enumerate}\addtocounter{enumi}{0} % 
 
  \item Shuffle the sequences randomly.

  \item Build a new order for the sequences by placing the clusters one after the other following their size (from the smallest to the biggest).

 \item Calculate the distance between consecutive sequences following the new order. Avoid calculating the distance in case of self-match. 
\end{enumerate}
After the Warm-up, each sequence has an approximate $nnd$.
This procedure resembles the original HOT SAX, since we are essentially calculating distances among sequences belonging to the same cluster. A real advantage can be obtained only if the approximate $nnd$s can be improved substantially at a low computational cost. 
%%%%%%%%%%%% NEW %%%%%%%%%%%
%%%%%%%%%%%% End NEW %%%%%%%

\subsection{Short range time topology}\label{ssec:timeTop}
%%%%%%%%%%%%%%%%%%%%% NEW %%%%%%%%%%%%%%%%%%%%%%%%%%%%%%%%
Here we detail how to exploit another property of the time series in order to improve the approximate $nnd$s obtained with the warm up procedure.
%%%%%%%%%%%%%%%%%%%%% END NEW %%%%%%%%%%%%%%%%%%%%%%%%%%%%

Time series display a form of time correlation first described as Consecutive Neighborhood Preserving (CNP) Property \citep{scrimp}. This property can be regarded as a form of time topology which makes possible to improve an approximate $nnd$ profile of a time series at a low cost  \citep{timeTopology}. 
This can be summarized by noticing that there is a high chance that the Euclidean nearest neighbors of two time-close sequences (i.e. starting at time $i$ and $i+1$) are also time-close, as in Fig.~\ref{fig:warmup} (right).
In practice, this can be seen as a form of autocorrelation of the $nnd$s which is particularly noticeable regarding the $ngh$ profile 
%%%%%%%% NEW %%%%%%%%%%%
where, often times, the following formula is true: 
\begin{equation}\label{eq:timeNeighbor}
ngh(i+1)=ngh(i)+1.
\end{equation}
%%%%%%% end new %%%%%%%%%
%
%
%%%%%%%%%% NEW %%%%%%%%%%%%%
This property can be used to improve the low quality $nnd$s obtained with the warm-up, where Eq. \ref{eq:timeNeighbor} is almost never verified.
%%%%%%%%%% end NEW %%%%%%%%%
%%%%%%%%%% NEW %%%%%%%
In fact, it is 
%%%%%%%%%% end new %%%
enough to check the following distance: $d\left(\frac{}{}i+1, ~ngh(i)+1 \right)$ for all of the sequences in order to obtain a better $nnd$ profile closer to the exact matrix profile. Since the time ordering is irrelevant when searching for the nearest neighbor, the opposite direction can also be employed: $d\left(\frac{}{} i-1, ~ngh(i)-1 \right)$. 
Also in this case, the number of distance calls is essentially identical to the size of the time series. 
In order to avoid useless calculations, before calculating the distance, it can be checked if it is already true that the neighbor of sequence at $i \pm 1$ is at  $ngh(i)\pm 1$. 
%
%
%%%%%%%%%%%%%%%%%%%%%%%%%%%%%%%% NEW %%%%%%%%%%%%%%%%%%%%%%%%%%%%%
This new $nnd$ profile is much better than the one resulting from the warm-up procedure.
We can now use the $nnd$s to reorder the loops for the discord search.  
%%%%%%%%%%%%%%%%%%%%%%%%%%%%%%%% END NEW %%%%%%%%%%%%%%%%%%%%%%%%%

\subsection{Rearranging the external loop}\label{ssec:external}

HOT SAX begins the external loop with likely discords, this is implemented by selecting at first those sequences that belong to small SAX clusters.

\subsubsection{Initial re-ordering process}\label{ssec:firstReorder}
For HST, we want to follow the same reasoning used in HOT SAX but enhance it. After the warm-up procedure and supplementing it with the time topology (Sec. \ref{ssec:timeTop}) all the sequences have an approximate $nnd$ value. 
%%%%%%%%%%% NEW %%%%%%%%%%%%%%
One is sure to make a reasonable guess putting at the beginning of the external loop those sequences which have a high $nnd$ value, and this choice is 
%%%%%%%%%%% END NEW %%%%%%%%%%
% 
likely better than just picking the smallest SAX clusters. % 
However, this $nnd$ profile is still rough, so it is better
%%%%%%%%%% new %%%%%%%
to smear it with a moving average.
%%%%%%%%%% end new %%%%
%
It has been shown, in fact, that the exact $nnd$  profile (matrix profile) displays a ``sort of smoothness'' \citep{scrimp, timeTopology}. A rough $nnd$ profile is likely to contain also spikes surrounded by dips. These spikes correspond  to ``false discord candidates''.
At variance, a true good discord candidate should belong to a ``peak'': it should be surrounded by quite a few other sequences with high $nnd$ values. This time coherence of the $nnd$s spans approximately the length of a sequence \citep{timeTopology}, for this reason the moving average takes into account $s+1$ sequences:
\begin{equation}\label{eq:moving}
 \overline{nnd}(i) =  \frac{1}{s+1}\sum_{j=-s/2}^{s/2}   nnd(i+j)  
\end{equation}
With this procedure, one gets rid of spikes not related to peaks.
At the borders, where the moving average cannot be done, we simply use the approximate $nnd$s. 
In summary, at the beginning of the external loop HST puts the sequences with the highest $\overline{nnd}$s, and at the end those with the lowest values. 
\subsubsection{Dynamic external loop order}\label{ssec:dynamicExternal}
%%%%%%%%%%%%%%%%%% NEW %%%%%%%%%%%%%%%%%%%%%%%%%%%%%%%%%%%%%%%%%%%%%%%%%%%
In HOT SAX, the external loop is fixed once for all. In HST the order of the external loop changes dynamically during the search.
The approximate $nnd$s, in fact, become smaller and smaller converging to the exact ones during the execution of the algorithm. 
%%%%%%%%%%%%%%%%%% NEW END %%%%%%%%%%%%%%%%%%%%%%%%%%%%%%%%%%%%%%%%%%%%%%%%
%
%%%%%%%%%%%%%%%%% NEW %%%%%%%%%%%%%%%%%%%%%%%%%
%%%%%%%%%%%%%%%%% END NEW %%%%%%%%%%%%%%%%%%%%%
%
As a result, those sequences that are more likely discords change during the execution of the code. It becomes reasonable to re-arrange the remaining parts of external loop in order to prompt high $nnd$ sequences at the beginning. 
This procedure takes into account the increasing information obtained as the algorithm proceeds. In this case, using a moving average does not produce good results since the quality of the approximation is becoming more accurate (a moving average would just diminish its precision).
Every time that a good discord candidate is found there is a distance call for (almost) all the sequences of the time series.  So it is likely that many $nnds$ have changed during the process.
For this reason, finding a good discord candidate is a natural point after which HST re-orders the remaining part of the external loop.

\subsection{Long range time topology}\label{ssec:longRange}
\begin{figure}[h!]
 \includegraphics[width=0.99\textwidth]{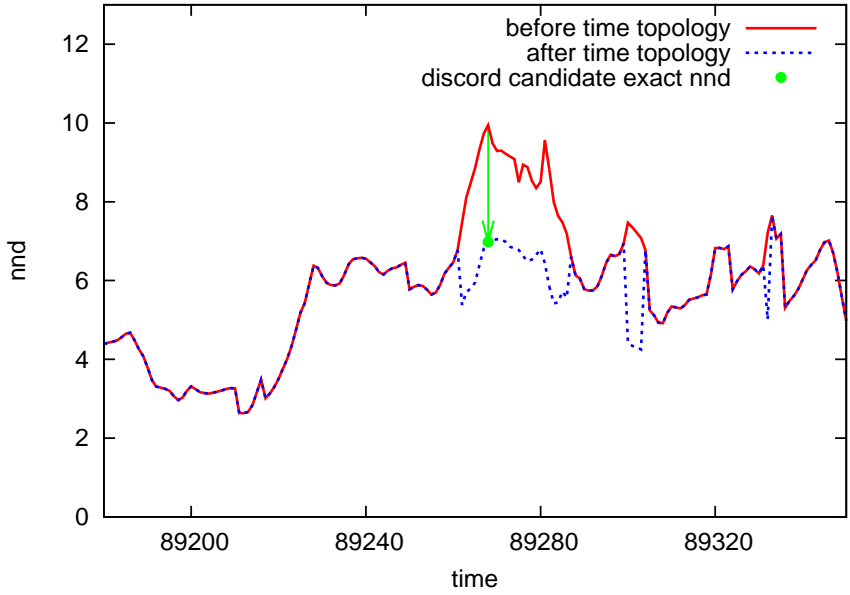}
 \caption{Sequence $89268$ has the highest approximate $nnd$ value ($9.94$) of a time series, and it is surrounded by a peak about $30$ sequences wide. It is a good discord candidate so its distance with all the other sequences (but self-matches) has been calculated ($\approx 2 \cdot 10^{5}$ calls). As a result its $nnd$ drops (green arrow) to the exact value $6.97$ (the green dot). The $nnd$s of its time neighbors (red curve) are not affected (since they are self-matches).  
 The long range time topology improves the approximate $nnd$s of many of the time neighbors of sequence $89268$ (blue dashed line) with a limited number of distance calls ($\le 2 \cdot s$)  .}\label{fig:longTime}
\end{figure}
%
% 
%%%%%%%%%%%% NEW %%%%%%%%%%%%%%%%%%%%%%%%%%
During the execution of  HST we make use of the CNT property in order to solve another problem.
Let's consider what happens when a good discord candidate is found during the search process. 
The ``pseudo-smoothness'' of the $nnd$ profile induced by the similarity between time-close sequences implies that a discord is not an isolated sequence but it belongs to a wide peak where time-close sequences show high $nnd$ values, so: 
\begin{itemize}
\item In the peak there are many sequences having a high $nnd$ value. For each of them the inner loop might require to scan most of the other sequences.
\item The width of the peak is not known.
\item The distance calls of a sequence do not contribute to the $nnd$s of its time-neighbors (they are self-matches). For this reason one needs to perform independent calculations for each of the sequences of the peak.
\end{itemize}
%%%%%%%%%%%% END NEW %%%%%%%%%%%%%%%%%%%%%%
%
%
The width of the peak (the number of sequences which belong to it), is likely proportional to the length of the sequences. This is a result of the non-self-match condition, since overlapping sequences do not contribute to each others $nnd$ and have similar properties. 
%
%
%
%A
% 
%
%
The time topology is very useful since it tells us that  
the Euclidean neighbors of: $i+1, i-1, i+2, i-2	, ...$  are likely time-close. This means that in many cases the following conditions are true: 
\begin{itemize}
 \item $ngh(i+1) = ngh(i)+1$
 \item $ngh(i-1) = ngh(i)-1$
 \item $ngh(i+2) = ngh(i)+2$
 \item \dots
\end{itemize}
With these suggestions, it is possible to make a very limited number of calls to the distance function and likely obtain much better $nnd$s for all of the sequences of the peak.
In practice, the \emph{long\_range\_time\_topology} functions essentially remove peaks as shown in  Fig. \ref{fig:longTime}. Notice that within the present algorithm one does not need to know the exact $nnd$s of the sequences of the peak. It is enough to find if these values are lower than the current highest exact one.
It is important for these functions to be well balanced, in order to avoid useless calculations. 
\begin{table}[h]
\begin{lstlisting}[caption={$Long\_range\_time\_topology\_forw$}]
for  j=1; j <= s; j++ do 
  if (nnd[i+j] < bestDist ) break// not a discord: check next one
  if (ngh[i+j] = ngh[i]+j ) return  // distance already calculated
  if (i+j      > N        ) return  // outside time series limits
  if (ngh[i]+j > N        ) return  // outside time series limits
  
  d = distance (i+j , ngh[i]+j)     // calculate distance 
 
  if (d < nnd[i+j] )  then 
    nnd[i+j] = d                    // update distance
    ngh[i+j] = ngh[i]+j             // update neighbor
  else 
    return //the time topology provides no improvement!            
  end if   
end for   
\end{lstlisting}
\end{table}
A function of this kind contains a loop running on the $s$ time-closest sequences of the one under consideration, exactly because they overlap (beginning at position $i$, see Listing 1). %
If  \verb|nnd[i+j]| $<$  \verb|bestDist| (the highest so far exact $nnd$ value) there is no reason to try to improve it, since we already know that sequence \verb|i+j| cannot be a discord.
Otherwise the algorithm checks the distance between  \verb|i+j| and \verb|ngh[i+j]|.

It can happen that:
 \verb| nnd[i+j] < d|$\left(\frac{}{}\right.$~\verb|i+j, ngh[i]+j|$\left.\right)$.
In this case, the time topology is losing coherence. It becomes reasonable to skip the remaining part of the loop. Going to longer ranges would simply result in useless distance calls, i.e. slowing down the execution of the algorithm. 
Notice that both time directions should be taken into account. These procedures are embedded in two functions: 
\begin{itemize}
\item \emph{Long\_range\_time\_topology\_forw} (see Listing~1)
\item \emph{Long\_range\_time\_topology\_back}.
\end{itemize}

\subsection{Pseudocode: details}\label{ssec:pseudo}
\begin{table}[ht!]
\begin{lstlisting}[caption={HST Pseudocode}]
nnd = 99999999.9          // initialize array with all the nnds 

SAX()                      
Warm-up() 
Short_range_time_topology()
Sort_External(order)      // fill the array with the new ordering    

bestDist= 0.0             // initialize the current max nnd-value   

for  j=1; j<= N; j++ do   // the external loop        
 
  i= order[j]             // "i" is the sequence processed now
  can_be_discord = true   // sequence "i" can be the discord
  
  Avoid_low_nnds (i, bestDist) // skip sequences with low nnd

  if (can_be_discord == true) then //is "i" still a canidate?   
    Current_cluster(i, bestDist)//minimization (HOT SAX inner loop) 
  endif
  
  if (can_be_discord == true) then //is "i" still a canidate?  
    Other_clusters(i, bestDist)//minimization (HOT SAX inner loop) 
  end if   

  Long_range_time_topology_forw (i, bestDist) // level peaks
  Long_range_time_topology_back (i, bestDist) // level peaks

  if (can_be_discord == true) then // i is a good discord candidate 
    Update(bestDist)           // update the current max nnd-value
    Sort_Remaining_Ext(order)  // re-sort the external loop
  end if
end for   //  external loop
\end{lstlisting}
\end{table}
Here we detail the functions of HST as described in Listing 2.
The following functions are called before the external loop:
\begin{enumerate}
\item \emph{SAX}(), this function clusters the sequences \citep{sax}.
\item The \emph{Warm-up}() procedure obtains approximate $nnd$s for most of the sequences, as explained in Sec. \ref{ssec:warmup}. The $nnd$s are stored in the homonym integer array \verb|nnd[]|, while the neighbors (of each sequence) are stored in the integer array \verb|ngh[]|.
\item The \emph{Short\_range\_time\_topology}() function improves the quality of the $nnd$s (Sec. \ref{ssec:timeTop}).
\item The \emph{Sort\_External}(\verb|order|)  function creates an integer array containing the order of the sequences of the external loop, from the highest $\overline{nnd}$  to the lowest. This new order is stored into the integer array \verb|order[]|.

\end{enumerate}
Notice that no exact $nnd$ value has been found before the beginning of the external loop (up to line 6 of Listing 2). The double precision variable  \verb|bestDist|  stores the current highest exact $nnd$. At the beginning, this quantity is initialized to $0$ (the external loop is a maximization procedure so we initialize it with the lowest possible value). As a result the first sequence of the external loop will become the first good discord candidate and its exact $nnd$ will turn into the next \verb|bestDist|.
The index \verb|j| denotes  the external loop and it runs from 1 to $N$ (all the possible sequences). The order of the sequences of the external loop  follows instead the index \verb|i=order[j]|. 
In principle every sequence can be the discord, so the flag \verb|can_be_discord| is set to \verb|true|.

The function \emph{Avoid\_low\_nnds}() checks if \verb|nnd[i]| is smaller than \verb|bestDist| (the highest so far exact $nnd$), in that case the flag \verb|can_be_discord| is set to \verb|false|.  

The minimization phase comprises of two functions, scanning different parts of the search space: 
  \begin{enumerate}
      \item The function \emph{Current\_cluster}() calculates the distance between sequence $i$ and all the other sequences belonging to the same SAX cluster (but for self-matches). If the approximate $nnd$ of sequence $i$ drops below the \verb|bestDist|, then sequence $i$ cannot be a discord, the function returns control to the main program, setting the flag \verb|can_be_discord| to \verb|false|.
      \item The function \emph{Other\_clusters}() calculates the distance between sequence $i$ and all the other sequences of the time series. It begins scanning  small clusters and then it moves to bigger ones (but skipping the cluster already checked in the function \emph{Current\_cluster}). At any point, if the approximate $nnd(i)$ becomes smaller than \verb|bestDist|, the function sets the flag \verb|can_be_discord| to \verb|false| and then returns. Instead, when this function goes through all the remaining sequences and the flag \verb|can_be_discord| is still \verb|true|, we are sure that \verb|nnd[i]| is exact. The minimization procedure associated to sequence \verb|i| ends with this function. So \verb|i| is a good discord candidate. 
\end{enumerate}
The former two functions are almost identical to those of the inner loop of a normal HOT SAX execution. In HST, however, before passing to the next sequence of the external loop there are a few more steps:
\begin{itemize}
      \item The functions  \emph{Long\_range\_time\_topology\_forw()} and  \emph{Long\_range\_time\_topology\_back}() improve the  $nnd$s  of the time neighbors  of the sequence $i$ (sequences beginning at $i+1$, $i-1$, $i+2$ ...) as explained in Sec. \ref{ssec:longRange}.

      \item If the flag \verb|can_be_discord| is still true at the end of the inner loop,  it means that $nnd(i)$  is exact and is the highest value so far.  It is thus necessary to update the best so far $nnd$ value (\verb|bestDist|), with the help the function \emph{Update}().

      \item The condition \verb|can_be_discord = true| after the minimization procedure implies that essentially all the sequences were involved in at least one distance call. It is now a good point where to re-order the array governing the external loop (\verb|order|) as per Sec. \ref{ssec:dynamicExternal}. 
      The remaining sequences of the external loop are re-arranged according to the values of their updated $nnd$s: from those with the highest $nnd$s to those with the lowest values.

  \end{itemize}

\section{Validation}\label{sec:validation}
The aim of this section is to provide comparisons with direct competitors of HST. Since HST has not yet been parallelized we will limit the comparisons only to serial algorithms. 

There are two major families of metrics.
The first one is related to the accuracy of the algorithm itself, i.e. to which extent the returned anomalies are discords (approximate algorithms do not return exact values). For example, this can be obtained as the overlap of the discords and the anomalies found.
HOT SAX Time, on the other hand, does not need these tests since it returns the exact discords.

The other usual metric is a  measure of the speed. 
Although one is interested in the amount of seconds required to find the anomaly, the actual calculation time can be masked by many factors (e.g. programming language, hardware, compilation flags...).
As a result, the speed is often calculated as the number of calls to the distance function between sequences \citep{hotsax} (lower number of calls is better). In the past, more than 99\% of the calculation time was spent within this function. However, if one uses the distance function based on the scalar product as suggested by \cite{matrixII} (which is faster than the explicit distance among z-normalized sequences), this condition is no longer true, in particular if the number of calls is small. Although the time spent in the distance function is still most of the total, other functions add a non-negligible contribution to the total calculation time. In these cases both the number of distance calls and the running times might be needed. 
In order to compare HST with other algorithms we use the ratio of the distance calls (D-speedup) or the ratio of the running times (T-speedup)  (see Sec. \ref{sec:background}). If the calculation times are long enough (at least tens of seconds) the two indicators are expected to be close (within 20\%-30\%). At variance, when the running times are short ($\approx 1s$), the T-speedup is not particularly a good indicator since it is affected by other functions that require tenths of seconds.

If the number of distance calls is either not available or not accurate, like in the case of SCAMP and DADD, we will consider only the execution times. Notice that the available implementations of these two algorithms are written in C/C++. 
%%%%%%%%%%%%%%%%%%%% NEW %%%%%%%%%%%%%%%%%%%%%%%%%%%%%%%%%%%%%%%%%%
HST instead, has been implemented in Fortran. This is not an important problem since 
these programming languages have very similar performances, as it can be seen for example at  \textit{The Computer Language Benchmarks Game}, \cite{benchmarkSite}. In the case of RRA the available implementation is written in JAVA, this can slow down the execution times and for this algorithm, the comparisons are limited to the number of distance calls.
%%%%%%%%%%%%%%%%%%%% END NEW %%%%%%%%%%%%%%%%%%%%%%%%%%%%%%%%%%%%%%
%
%%%%%%%%%%%%%%%%%%%% NEW %%%%%%%%%%%%%%%%%%%%%%%%%%%%%%%%%%%%%%%%%%
The datasets used to validate RRA  \citep{senin} \citep{seninVali} are a heterogeneous collection of time series often used in literature, ranging from ECG to breathing time series. Their length goes from few thousand to a few hundred thousand points. For this reason, we also used them in the present article.
Some of these time series are extracts of longer datasets, for example those of the Physionet repository \citep{mit-cuore} \citep{physio}. In order to provide a uniform comparison with RRA we stick with the ones already used for its validation.
%%%%%%%%%%%%%%% new %%%%%%%%%%
We included a real dataset containing more than $10^8$ points  \citep{insect}.
%%%%%%%%%%%%%%% end new %%%%%%%%
%
We also used synthetic time series since they allow us to have full control of their properties. 

It is important to say that all these algorithms (but SCAMP) include pseudo random processes, so the number of distance calls for a given dataset fluctuates. Determining the speed of the algorithm with a single test might be misleading.
We overcome this problem by averaging 10 runs on each dataset.
%%%%%%%%%%%%%%%%%%%% END NEW %%%%%%%%%%%%%%%%%%%%%%%%%%%%%%%%%%%%%%

%\subsection{HST, HOT SAX and RRA}\label{ssec:firstComparison}

\subsection{HOT SAX and HST}\label{ssec:hstHS}
\begin{table}[!t]
 \centering
 \begin{tabular}{l|cr|rr|r|c} 
              &                       &                   &    \multicolumn{2}{|c|}{ \# of distance calls}                               &               &    Runtimes [s]  \\
  file        &  s,  P, alphabet      &   length          &     HOT SAX         &     HST    &     D-speedup   &  HST      \\  
  \hline 
 Daily commute       &  345, 15, 4    &          17 175   &   819 802      &   260 615  &   3.14             & $0.18$   \\   
 Dutch Power         &  750,  6, 3    &          35 040   &  3 428 728     &   259 820  &  13.19             & $0.32$   \\  
  ECG 0606           &  120, 4, 4     &           2 299   &   20 621       &     8 166  &   2.52             & $0.017$   \\  
  ECG 308            &  300, 4, 4     &           5 400   &   149 329      &    25 959  &   5.75             & $0.039$   \\  
 ECG 15              &  300, 4, 4     &          15 000   &    215 928     &    91 970  &   2.35             & $0.088$   \\  
 ECG 108             &  300, 4, 4     &          21 600   &   1 456 777    &   106 737  &  13.65             & $0.22$   \\  
 ECG 300             &  300, 4, 4     &         536 976   &   46 382 574   &  6 547 211 &   7.08             & $4.18$   \\ 
 ECG 318             &  300, 4, 4     &         586 086   &    46 827 423  &  4 426 685 &  10.58             & $3.21$   \\  
  NPRS 43            &  128, 4, 4     &           4 000   &   79 340       &    35 466  &   2.23             & $0.02$   \\  
 NPRS 44             &  128, 4, 4     &          24 125   &   398 471      &   136 658  &   2.91             & $0.10$  \\  
 Video               &  150, 5, 3     &          11 251   &   210 089      &    91 397  &   2.30             & $0.056$   \\  
  Shuttle, TEK 14    &  128, 4, 4     &            5 000  &  490 342       &    65 353  &   7.50             & $0.06$   \\ 
  Shuttle, TEK 16    &  128, 4, 4     &            5 000  &  546 369       &    69 912  &   7.81             & $0.055$   \\ 
  Shuttle, TEK 17    &  128, 4, 4     &            5 000  &  476 616       &    71 436  &   6.67             & $0.057$    \\ 
 \end{tabular} 
\caption{The first column contains the names of the datasets, then we include the lengths of the sequences (s), the SAX parameters (P, alphabet) and the length of the time series. The number of distance calls for the first discord for HOT SAX and HST (lower is better) are in the columns with the names of the two algorithms. The D-speedup shows that in all the cases under observation HST is at least two times faster than HOT SAX, on four occasions it is more than 5 times faster and for three datasets it is more than 9 times faster. The last column shows the running times for HST in seconds.}  
 \label{tab:hsVhs}
\end{table}
%
%%%%%%%%%%%%%% NEW %%%%%%%%%%%%%%%%%%%%%%%%%%
At the beginning, we compare HOT SAX and HST.
%%%%%%%%%%%%%% END NEW %%%%%%%%%%%%%%%%%%%%%%
The reference HOT SAX code (written by us) has many similarities with its HST counterpart, it is written in the same language and many of the subroutines are essentially identical.
%T
We consider the same datasets and conditions that were used to compare RRA and HOT SAX (Tab. \ref{tab:hsVhs}).
In 8 cases out of 14, HST is more than 5 times faster than HOT SAX, reaching D-speedup peaks of 13 for two datasets.
\begin{table}
 \centering
 \begin{tabular}{c|rr|r||rr|r} 
                     &    \multicolumn{2}{c|}{ \# of distance calls} &           &       \multicolumn{2}{c|}{ Runtimes [s]}   &               \\
  file               &    HOT SAX      &   HST            &  D-speedup &     HOT SAX               &        HST      &   T-speedup   \\ 
  \hline 
 Daily commute       &    4 373 481  &     819 880       &   5.33  &     1.78   &    0.45                 &   3.97   \\   
 Dutch Power         &   20 326 437  &   1 043 572       &  19.48  &    14.40   &    0.94                 &   15.29  \\   
 ECG 15              &   10 947 552  &     705 152       &  15.53  &     3.64   &    0.30                 &   12.26  \\  
 ECG 108             &   10 194 725  &     856 132       &  11.91  &     4.07   &    0.73                 &   5.59   \\  
 ECG 300             &  447 184 547  &  44 697 489       &  10.00  &   147.49   &   17.14                 &   8.60   \\ 
 ECG 318             &  269 580 847  &  37 740 624       &   7.14  &    90.99   &   14.54                 &   6.26   \\  
  NPRS 43            &    1 005 254  &     187 478       &   5.36  &     0.20   &   0.056                 &   3.64   \\  
 NPRS 44             &    6 748 679  &   1 666 487       &   4.05  &     1.13   &    0.45                 &   2.52   \\  
 Video               &    2 742 811  &     481 800       &   5.69  &     0.62   &    0.15                 &   4.05   \\   
  Shuttle, TEK 14    &    1 500 550  &     265 364       &   5.65  &     0.34   &    0.086                &   3.98   \\ 
  Shuttle, TEK 16    &    1 613 129  &     274 172       &   5.88  &     0.38   &    0.095                &   3.98   \\ 
  Shuttle, TEK 17    &    1 460 009  &     276 351       &   5.28  &     0.33   &    0.096                &   3.50   \\ 
 \end{tabular} 
\caption{The number of distance calls and running times for the first 10 discords for HOT SAX and HST.}  
 \label{tab:tempi10D}
\end{table}
% 
% %
%
%
%%%%%%%%%%%%%%%%%%%%%%%%%%%% NEW %%%%%%%%%%%%%%%%%%%%%%%%%%%%%%
For these calculations, it is not particularly meaningful to consider the T-speedup. The reason is that the execution times for the first discords are very short.
%%%%%%%%%%%%%%%%%%%%%%%%%%%% END NEW %%%%%%%%%%%%%%%%%%%%%%%%%%
As a result, an important percentage of the running time is spent in functions other than the distance one. 
We keep these tests since they have the same structure as those used for the evaluation of RRA \cite{seninVali}.
In order to obtain conditions that represent better the speed difference between the two algorithms, we expand the calculation over 10 discords. This is legitimate since both HOT SAX and HST report exact values. 
Notice that the ``non self-match condition'' implies that the total number of discords in a dataset is limited. At most, there can be $(N/s)+1$ discords. For these reasons, we excluded ECG 308 and ECG 0606.  
In Tab. \ref{tab:tempi10D} we can see that HST is substantially faster than HOT SAX.  The runtimes for HST are still short and so  there is an important impact from other functions.
Only for the longer time series (EGG 300 and ECG 318), HST requires more than 1 second for calculating the first 10 discords. In those cases, the D-speedup and T-speedup are within 20\% of each other, while for shorter calculations the gap can be wider. 
%%%%%%%%%%%% new %%%%%%%%5
Although for quick calculations the T-speedup is not very accurate we can use the D-speedup as a guide to better understand the kind of problems where HST is faster. When the task becomes more demanding (10 discords instead of 1), HST becomes from 4 to 19 times faster than HOT SAX, in terms of distance calls and 3 to 15 times faster in terms of runtimes. At this point, one might be interested in understanding the characteristics of the discord search which determine this variety of results.
%%%%%%%%%%%% end new %%%%%%%%

\subsection{The complexity of a search: cost per sequence}\label{ssec:cps}

%Datasets:
%https://www.cs.ucr.edu/~eamonn/discords/
%%%%%%%%%%%%%%%%%% NEW %%%%%%%%%%%%%%%%%%%%%%%%%%%%%%%%%%%%%%%%%%%%%%%%%%%%%%%%%%%%%%%%%%%%
In this section, we provide an analysis that will lead to the characterization of complex searches. We need an indicator that allows us to compare searches on time series of different lengths and according to the different parameters of the search.
%%%%%%%%%%%%%%%%%% END NEW %%%%%%%%%%%%%%%%%%%%%%%%%%%%%%%%%%%%%%%%%%%%%%%%%%%%%%%%%%%%%%%%
There are two main approaches regarding algorithms and the problems to be solved:
\begin{itemize}
 \item One can study the asymptotic complexity of the algorithm, as a function of some parameters of the problem, for example the size of the dataset. This approach is particularly useful when the algorithms do not depend on the data to be analyzed (like the Matrix Profile ones), since it allows to precisely compare different algorithms.
 
 \item If the execution is sensitive %in respect 
 to the specific dataset under investigation, one can use the execution properties (time, space required,...) to
 characterize specific instances of the problem.
% 
%%%%%%%%%%% NEW %%%%%%%%%
For example, 
%%%%%%%%%%% end NEW %%%%%%%%%
 one can order the problems in terms of difficulty.
\end{itemize}
In this section, we take the second point of view since we want to characterize the problems.
% defined by:
%%%%%%%%%% NEW %%%%%%%%
A discord search is influenced by:
%%%%%%%%%% end NEW %%%%
\begin{itemize}
 \item the content of the time series under investigation.
 \item the length of the time series $N$.
 \item the length of the sequences $s$.
 \item the number of discords to be found.
\end{itemize}
%
% 
%
%
%%%%%%%% NEW %%%%%%%%%
In the following, we will define an indicator to measure the difficulty of discord searches. As a benchmark, we will use the results of HOT SAX, since it is popular, easy to implement, and it is not optimized on a specific kind of time series. Moreover, both HST and HOT SAX are based on SAX, so it is safe to compare two searches when the SAX parameters are identical.
%%%%%%%% END NEW %%%%%
%
%
%
%
%%%%%%%% NEW %%%%%%%%%%%%%
For some searches, HOT SAX already represents a valid solution, while we are interested in understating in which instances HST is much better. 
%%%%%%%% END NEW %%%%%%%%%

%

Since the number of distance calls %depends on 
%%%%%% new %%%%%%%
grows with
%%%%%% end new %%%
 the length of the time series, one might think that also the speedup between HST and HOT SAX might grow with $N$.
The experiments, however, do not confirm this hypothesis. In Tab. \ref{tab:hsVhs}, the best results of HST have been obtained with ECG 108 whose length is $21600$ (speedup of $13$ in respect to HOT SAX). At variance ECG 300 is more than 24 times longer than ECG 108 but the speedup is about~$7$. 
Since the length of the time series is not the main parameter determining the speedup, we will now try to find other quantities with a more clear impact. %
The performance difference is likely related to both the structure of the signal and the  length of the sequences.
%%%%%%%%%%%%%%%%   NEW %%%%%%%%%%%%%%%%
The total number of distance calls naturally grows with the length of the time series since each sequence must be checked. For this reason, it is meaningless to rank the discord searches of two time series having very different lengths, for example one containing $10^6$ points and the other $10^8$ points. However, some discord searches are clearly easier than others (even if the number of points is big). For example, if one time series is constant but for a single bump, even a person at first glance  can identify the anomaly with the bump. On the opposite side, there are discord searches on ``short'' time series which require lengthy computations even on fast machines. 
%%%%%%%%%%%%%%%% END NEW %%%%%%%%%%%%%%
%
For these reasons, we define the \textit{cost per sequence} (\textit{cps}) as the number of distance calls needed to find the first k-discords divided by the total number of sequences $N$ and by the number of discords~$k$:
$$
\mbox{cost per sequence} = cps  = \frac{\mbox{\# of distance calls}}{Nk}.
$$
The $cps$ can be used to compare searches on time series of different lengths and to order them in terms of difficulty. 
%
%
%%%%%%%%%%%%%%%% NEW %%%%%%%%%%%%%%%
In general, there are two complementary interpretations of the \textit{cps}:
%%%%%%%%%%%%%%%% END NEW %%%%%%%%%%%
%
\begin{itemize}
\item It is the average number of distance calls per sequence  %needed to solve the search 
%%%%%%%%%%% NEW %%%%%%%%%%%
and per discord. %W
%%%%%%%%%%% end NEW %%%%%%%
%
\item %%%%%%%%%%%%%%%%% NEW %%%%%%%%%%%
If the exact $nnd$ profile (matrix profile) has a few big ``bumps'', and if it is possible to identify with little computational cost (1 distance call each) those sequences which are ``normal'' (the background with low $nnd$ values), then the \textit{cps} 
      %%%%%%%%%%%%%%%%% END new %%%%%%%
%It 
can be considered as a hint regarding the quantity of ``good discord candidates'' (bumps) encountered during the search process, since each them requires about $N$ distance calls. 
%%%%%%%%%%%%% NEW %%%%%%%%%%%%
%
%%%%%%%%%%%%% end NEW %%%%%%%%
\end{itemize}
%
%A
%
%%%%%%%%%%%%%%%%%%%% NEW %%%%%%%%%%%%%%%%%%%%%%%%%%%%%
Let's recall that HOT SAX is based on a ``magic'' ordering of the loops which allows one to skip most of the calls of a brute force approach. 
From this perspective, the \textit{cps} can be seen as the inverse of the ``magic'': a low $cps$ is an indication that the ``magic'' ordering is working well, while if the $cps$ is high, more distance calls are needed and  the ``magic'' is less effective. Let's consider two opposite cases. If one has access to a ``perfect magic'' all the sequences which are not the discord require just one distance call to a close neighbor to be discarded, while only for the discord it is required to calculate the distance with all of the $N-1$ sequences. In this case, the total number of calls is $2(N-1)$, and the \textit{cps} $\approx 2$.  
In an unfortunate event where the ``magic'' ordering fails completely, one essentially re-obtains a brute force approach that returns the maximum possible \textit{cps} $\approx N$. 

Although the \textit{cps} is a rather intuitive concept, as far as the authors know, it has never been formalized or explicitly used to classify the complexity of discord searches.

We will now use HOT SAX \textit{cps} for defining complex discord searches. 
%%%%%%%%%%%%%%%%%%%% END NEW %%%%%%%%%%%%%%%%%%%%%%%%%
%
%
\begin{table}[t!]
\centering
 \begin{tabular}{l|cc|r}
                  & \multicolumn{2}{c|}{\small{Cost per sequence}} & \\
  file                & HS  & HST & D-speedup    \\
  \hline   
 ECG~0606          &      9   &    4  &   2.52     \\ 
 ECG~15            &     14   &    6  &   2.35     \\
 NPRS~44           &     16   &    6  &   2.91     \\
 Video             &     19   &    8  &   2.30     \\ 
 NPRS~43           &     20   &    9  &   2.23     \\
 \hline
 ECG~308           &     28   &    5  &   5.75     \\
 Daily~Com.        &     48   &   15  &   3.14     \\
 \hline
 ECG~108           &     67   &    5  &  13.65     \\
 ECG~318           &     80   &    8  &  10.58     \\
 ECG~300           &     87   &   12  &   7.08     \\
Shuttle,~TEK~17    &     95   &   14  &   6.67     \\            
 Dutch~Power       &     98   &    7  &  13.19     \\
Shuttle,~TEK~14    &     98   &   13  &   7.50     \\
Shuttle,~TEK~16    &    109   &   14  &   7.81       
 \end{tabular}
 \caption{The second column shows the ``cost per sequence'' ($k=1$) expressed as the number of distance calls over the time series length (for HOT SAX) rounded to the first integer value. The third column contains the same quantity but referred to HST. The fourth column contains the D-speedup. In this table, the files are ordered according to an increasing HOT SAX cost per sequence.}
 \label{tab:callXseq}
\end{table}
%
%
%
%%%%%%%%%%%% new %%%%%%%%%%%%%%
It is possible to notice that HOT SAX \textit{cps} can exhibit strong variations, while HST \textit{cps} is more stable.
%%%%%%%%%%%% end new %%%%%%%%%%
For better visualizing this fact we set at the beginning of Tab. \ref{tab:callXseq} those datasets where HOT SAX performs better, and at the end those which require more calculations.    
%H
%%%%%%%%%%%%%%% new %%%%%%%%%%%%
For these problems, HST \textit{cps} is in between 4 and 15.
%%%%%%%%%%%%%%% end new %%%%%%%%
Notice that the \textit{Warm-up()} and \textit{Short\_range\_time\_topology()} procedures already account for about 2 distance calls per sequence. Even in the case in which the first sequence to be analysed turns out to be the discord,  HST should require at 
%%%%%%%%%%%% new %%%%%%%%%
least \textit{cps}$\sim 3$. 
%%%%%%%%%%%% end new %%%%%
%$. 
For less ``complex searches''  (those for which HOT SAX needs less than 20 distance calls per sequence), the maximum speedup obtainable by HST is limited by the structure of the algorithm itself. For example, HOT SAX requires just $9$ distance calls per sequence for ECG 0606. The maximum theoretical D-speedup attainable in this case by HST is $\sim 3$; while the actual D-speedup is $2.52$.  In fact, we can notice that for none of the time series with low \textit{cps} (less than 20) the D-speedup is higher than 3. While for all the sequences with a cost per sequence equal to or higher than 67 the D-speedup is greater than 6, reaching peaks of 13 in two cases.

\begin{figure}[h!]
  \includegraphics[width=0.499\textwidth]{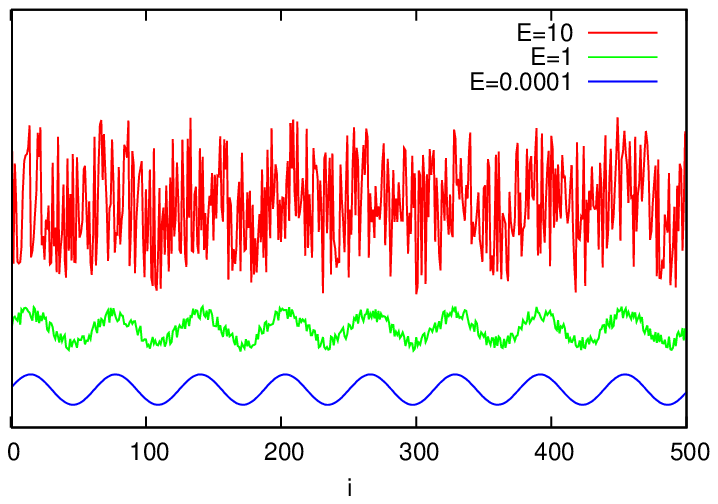}
  \includegraphics[width=0.499\textwidth]{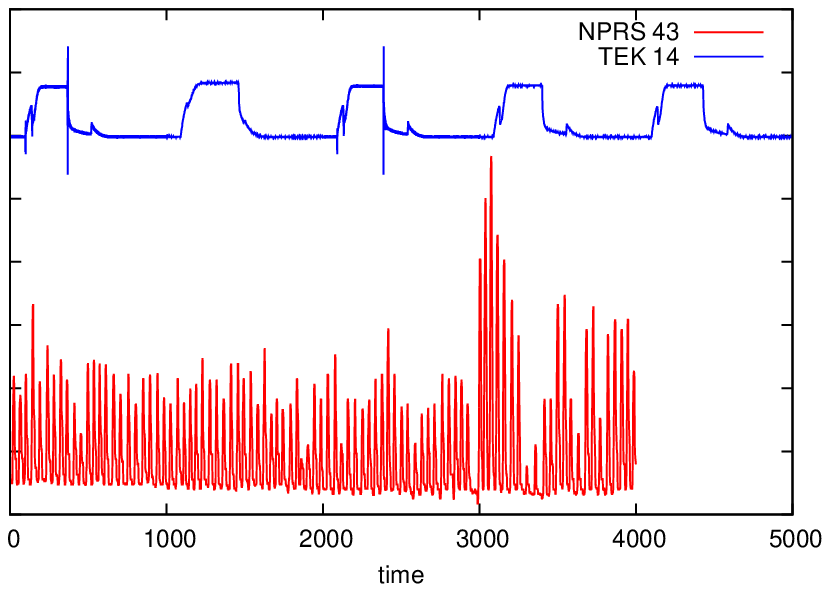}
 \caption{
  %(
 (Left) Extracts of synthetic time series with different levels of noise $E$. In order to avoid overlaps, they have been shifted vertically. 
 (Right) NPRS 43 and TEK 14 time series in arbitrary units. TEK 14 looks simpler and smoother, but it has a higher cost per sequence than NPRS 43.} \label{fig:noiseSpeedup}
\end{figure}
\begin{table}[h!]
\begin{minipage}{0.5\linewidth}
\raggedleft
 \begin{tabular}{c|cc|cc}
% \begin{tabular}{p{5mm}p{15mm}p{12mm}p{7mm}p{7mm}}
         &   \multicolumn{2}{|c|}{\small{\# of distance calls}}  &   \multicolumn{2}{c}{\small{cps}} \\  
   E     &   HOT SAX         &    HST      &    HS       & HST   \\
   \hline
 0.0001  &     24 527 170     &   234 707  &   1 226      &    12     \\  
 0.001   &     19 560 251     &   329 397  &     978      &    16     \\
 0.01    &      5 183 885     &   313 363  &    259       &    16       \\
 0.1     &      1 912 774     &   207 881  &     96       &    10       \\
 0.5     &      1 331 203     &   165 142  &     67       &     8      \\
 1.0     &      1 564 755     &   219 777  &     78       &    11       \\
 5.0     &      3 310 974     &   685 889  &    165       &    34     \\
 10.0    &     20 395 837     &  3 105 995 &   1 020      &   155 
 \end{tabular} 
\caption{ Number of distance calls and \textit{cps} (for one discord, $k=1$) as a function of the noise amplitude $E$ for the two algorithms, HOT SAX and HST.}
\label{tab:noise}
\vfill
\end{minipage}
\begin{minipage}{0.5\linewidth}
~~~                              % TRUCCO per separare TABELLA E IMMAGINE: spazio inter
  \includegraphics[width=0.95\textwidth]{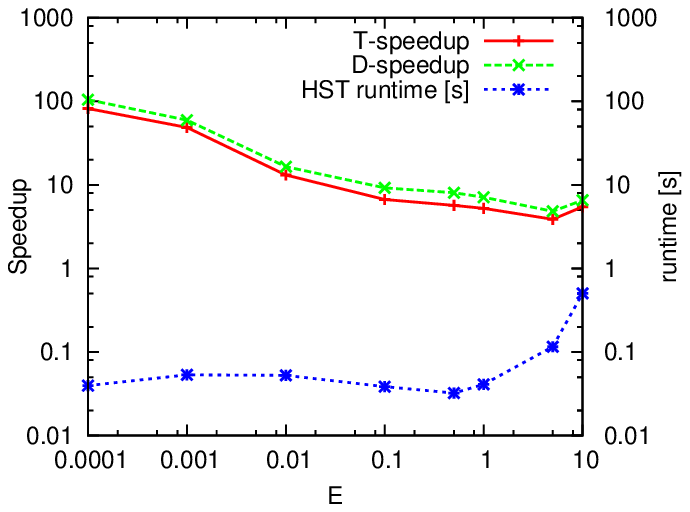}
  \captionof{figure}{noise}
  \captionsetup{width=.8\linewidth}
  \caption{The T-speedup and D-speedup of HST as a function of the noise amplitude $E$ in the case of a synthetic time series. All the time series comprise 20 000 points, $s=120$, $P=4$, and alphabet=$4$.}
  \label{fig:noise}
\end{minipage}
% %\end{center}
\end{table}
\subsubsection{Cost per sequence and "easy-looking" time series}\label{sssec:costEasyLooking}
It is interesting to notice that HOT SAX requires a lot of distance calls on many time series which look ``easier'' from a human point of view. For example the Shuttle Marotta Valve time series, TEK14, TEK 16, and TEK 17, appear simpler if compared with the respiration time series NPRS 43 and NPRS 44. As a comparison, we report NPRS 43 and TEK 14 in Fig. \ref{tab:noise} (right) since they have approximately the same length: TEK 14  shows 5 main patterns, while the interpretation of NPRS 43 is less easy.  
We can further investigate this anti-intuitive behavior with the help of synthetic time series. In the following, we consider a very simple function consisting of a re-scaled sine function plus a uniform random noise:
%(following Eq.~\ref{eq:sin}).
%%%%%%%%%%%%%%%%%%%% NEW %%%%%%%%%%%%%%%%%%%%%%%%%%%%%
 \begin{equation}\label{eq:sin}
  p_i =   \frac{  \sin(0.1 \cdot  i) + E\epsilon + 1} {2.5}
 \end{equation}
%%%%%%%%%%%%%%%%%%%% END NEW %%%%%%%%%%%%%%%%%%%%%%%%%
%
By changing the amplitude $E$ of the random noise, $\epsilon\sim U(0,1)$,  one can obtain a smoother or more ragged time series (Fig. \ref{fig:noiseSpeedup}, left).
We found that HST performs much better than HOT SAX when the noise is very low. When the amplitude of the noise $E$ is $0.0001$, HST is about 104 times faster than HOT SAX (see Fig. \ref{fig:noise}), in terms of distance calls (and 82 times faster in terms of runtimes). 
In particular, this is due to the fact that the HST performances are not so much affected by low noise, (Tab. \ref{tab:noise}), while HOT SAX degrades significantly.
In the cases of very low \textit{noise}/\textit{signal} ratios, HOT SAX $cps$ exceeds 1200, while it is about 12 for HST.
On the other side of the spectrum, when the noise is much higher than the signal, both HOT SAX and HST performances degrade (but HST keeps on being much faster).
%
%
%

% 
%As. 
%%%%%%%%%%%% NEW %%%%%%%%%%%%
We make here an educated guess on the reason why HOT SAX struggles to analyse ``apparently easy'' time series.
%%%%%%%%%%%% end new %%%%%%%%
%
When there are many similar sequences and patterns, the matrix profile is likely to have a large number of peaks  with very similar heights. For this reason, HOT SAX needs a lot of calculations to disambiguate which of them is the highest one, while HST is quicker due to its better ``aiming mechanism''.

HST is still an excellent choice when the noise is as high as the signal itself,  while it is less effective when the noise level is much higher than the signal. This is not a surprise: when the noise is 10 times higher than the signal, it dominates the time series. As a result, the time topology (linked to autocorrelation) ceases to be helpful in finding discords. An anomaly search, %discord search
 however, is useful when most of the sequences are expected to be ``normal'', and this condition is not satisfied when the noise is 10 times higher than the signal itself.

 As far as the authors know these are the first results regarding an exact discord algorithm capable of being more than two orders of magnitude faster than HOT SAX.

\subsubsection{Cost per sequence and length of the discords}\label{sssec:cpsLength}
An important challenge regarding anomaly detection is related to the increasing length of time series. This can be due to the longer data collection times, but also to improved accuracy of the sensors that are able to sample more frequently. The same phenomenon can thus be described with a larger number of points. It becomes desirable to understand to which extent the calculation speed is affected by the length of the discords.  
Of course, if we increase the length of the sequences we can expect that the total execution time increases. This is a direct consequence of the fact that most of the time is spent calculating distances, and the execution time of a single distance call is approximately proportional to the length of the sequences. 
On the other hand, there is no obvious reason why the number of distance calls should increase. Actually, the search space shrinks as $s$ increases  because of the non self-match condition. According to our model, the cost per sequence is connected with the peaks of the $nnd$ profile. A large peak includes many sequences with high $nnd$ value which are difficult to analyse from HOT SAX.
The width of a peak, depending on the non self-match condition, should grow with the length of the sequences, and consequently also the number of good discord candidates.
Since the long range time topology procedure helps in leveling the peaks, HST should suffer this problem less than HOT SAX. As a result, one can expect high speedups when searching long discords. 
We tested this fact with the two longest datasets of Tab. \ref{tab:hsVhs}, ECG 300 and ECG 318, by increasing the length of the sequences $s$.
\begin{table}[h!]
\begin{center}
\begin{tabular}{c|rr|c}
           \multicolumn{4}{c}{\small{ECG 300}}      \\
          &  \multicolumn{2}{c|}{\small{Cost per sequence}}          &   \\ 
   $s$    &       HOT SAX    &     HST   &             D-speedup \\    
   \hline
   300          &   87     &        12      &   7       \\
  460          &   201     &        11       &  17       \\
  920          &   494     &        10       &  50       \\
 1380          &  1 553     &       19      &  82       \\
 1880          &   857     &       10       &  83       \\
 2340          &   750     &       10       &  71 
 %
 %
 %536976
 % 
\end{tabular}
%
%\begin{center}
\begin{tabular}{||c|rr|c}
   %\hline
           \multicolumn{4}{c}{\small{ECG 318}}   \\ 
          &  \multicolumn{2}{c|}{\small{Cost per sequence}}          &   \\            
   $s$    &       HOT SAX    &     HST   &             D-speedup \\    
   \hline
   300          &        80     &         7       &     11      \\
   460          &       113     &         6       &     18      \\
   920          &       510     &         9       &     56      \\
  1380          &       703     &        12       &     59      \\
  1880          &     2 026     &        21       &     94      \\
  2340          &     3 137     &        31       &    101 
\end{tabular}  
\caption{Cost per sequence and speedup as a function of the length of the sequences $s$ for HOT SAX and HST, for the dataset ECG 300 (left) and for ECG 318 (right). The other SAX parameters are $P=4$, and alphabet=4.}
%\label{tab:300lungo}
%
\label{tab:318lungo}
\end{center}
\end{table}
The values of Tab. \ref{tab:318lungo} confirm that the cost per sequence for HOT SAX is strongly dependent on the length of the sequences. Even in the case of sequences of length 920, the speedup can exceed 50, reaching peaks of more than 100 for longer discords. This is another example where HST can outperform HOT SAX by more than 2 orders of magnitude, moreover this specific condition (longer sequences) is expected to be more relevant in the future.

In summary, by using the \textit{cps} we gained insights regarding problems that are particularly difficult to treat with HOT SAX, and can greatly benefit from using HST. In those cases, HST can be more than 2 orders of magnitude faster than HOT SAX.
In particular, searches related to long discords become more complex. This is due to the fact that one should not look at the time series but rather at the structure of the search space, i.e. the set of all the sequences, which has an important dependence in respect to the length of the sequences.

\subsection{RRA and HST}\label{ssec:hstRRA}
%
%
%%%%%%%%%%%%%%%%%%%%%% NEW %%%%%%%%%%%%%%%%%%%%%%%%%%
RRA is freely available on the internet \citep{senin} \citep{seninVali}, it is fast and so it represents an ideal comparison test for HST.
%%%%%%%%%%%%%%%%%%%%%% END NEW %%%%%%%%%%%%%%%%%%%%%%
%
% 
%
%
RRA finds anomalies by exploiting the Kolmogorov complexity of the SAX words. At variance in respect to HOT SAX, the length of the discord is not required as an input parameter but it is obtained as a result of the calculation. The anomalies found with RRA, however, do not coincide with discords. Although there is a good overlap between the first discord of a time series and the anomaly found with RRA, the probability that they coincide is less than $1$. Multiple applications of the algorithm in order to find the k-th discord, provide less and less correspondence with the exact solutions. For this reason, at the time of the comparison between HST and RRA we will limit the calculations to the first discord. It is also interesting to notice that the SAX parameters affect only the speed of the algorithm in the case of HST, while they modify also the position of RRA anomalies.

\begin{table}[ht!]       
\begin{centering}
 \begin{tabular}{l|cr|rr|r}
              &                       &                   &     \multicolumn{2}{|c|}{\small{\# of distance calls}}   &                  \\
  file        &  s,  P, alphabet      &   length          &     RRA          &     HST    &     D-speedup    \\
  \hline
 Daily commute       &  345, 15, 4    &          17 175   &    388 504       &   260 615  &  1.49   \\  
 Dutch Power         &  750,  6, 3    &          35 040   &    1 801 971     &   259 820  &  6.93   \\ 
  ECG 0606           &  120, 4, 4     &           2 299   &     35 464       &     8 166  &  4.34   \\ 
  ECG 308            &  300, 4, 4     &           5 400   &    101 850       &    25 959  &  3.92   \\ 
 ECG 15              &  300, 4, 4     &          15 000   &    352 331       &    91 970  &  3.83   \\ 
 ECG 108             &  300, 4, 4     &          21 600   &    532 476       &   106 737  &  4.99   \\ 
 ECG 300             &  300, 4, 4     &         536 976   &    199 865 375   &  6 547 211 &  30.52   \\
 ECG 318             &  300, 4, 4     &         586 086   &    58 462 005    &  4 426 685 &  13.2   \\ 
  NPRS 43            &  128, 4, 4     &           4 000   &      89 620      &    35 466  &  2.52   \\ 
 NPRS 44             &  128, 4, 4     &          24 125   &      438 957     &   136 658  &  3.21   \\ 
 Video               &  150, 5, 3     &          11 251   &      165 758     &    91 397  &  1.81   \\  
  Shuttle, TEK 14    &  128, 4, 4     &            5 000  &    326 981       &    65 353  &  5.00   \\
  Shuttle, TEK 16    &  128, 4, 4     &            5 000  &    341 405       &    69 912  &  4.88   \\
  Shuttle, TEK 17    &  128, 4, 4     &            5 000  &    417 860       &    71 436  &  5.84   \\
 \end{tabular}
 \caption{Details of the calculations: ``s'' is the length of the sequences (also referred to as ``window''), ``P'' is the number of letters in which the sequence is divided, while the total number of available letters is the parameter ``alphabet''. ``Length'' refers to the number of points in the time series under observation. The values of s, P, and alphabet follow those of \citep{seninVali}. The minor differences are due to the fact that,  for our algorithm, P must divide exactly s (while this is not the case for the algorithm implemented in Grammarviz 3.0). Columns 3 and 4 show the average number of distance calls needed by RRA and HST for finding the first discord. The number of distance calls has been averaged over 10 runs for each dataset and algorithm. The distance-speedup (or D-speedup) has been calculated as the ratio between the number of distance calls of RRA over those of HST.
All these datasets can be downloaded from GitHub \citep{gram2site}.}\label{tab:rraVhs}
\end{centering}
\end{table}
The results reported for RRA \citep{seninVali} \citep{gram2site} are substantially different in respect to the numerical experiments we obtained with their code (Tab. \ref{tab:rraVhs}). This might be due to the choice of the \verb|--strategy| parameter (the available choices are \verb|NONE|, \verb|EXACT|, and \verb|MINDIST|). This parameter is used to further reduce the search space and it can improve significantly the speed of the algorithm.
For some datasets, one can obtain essentially the same anomaly with different flags.  
In many cases, however, the precision loss is not acceptable. As an example, the position of the exact discord of length $128$ for the time series TEK 14 is $3852$. The position of the RRA anomaly with \verb|--strategy EXACT| is $4717$, with \verb|--strategy MINDIST| is $4320$ (this time series contains $5000$ points, as explained in Tab. \ref{tab:rraVhs}).
Since a user cannot know in advance if one strategy provides enough accuracy, a valid comparison should be limited to the execution flag \verb|--strategy NONE|. Other strategies can provide faster executions but one needs to check afterward if the result is comparable with the discord (thus nullifying the improved speed). 
\begin{table}[h!]
\begin{lstlisting}[caption={RRA script for TEK 14}]
java -cp "target/grammarviz2-0.0.1-SNAPSHOT-jar-with-dependencies.jar" net.seninp.grammarviz.GrammarVizAnomaly -alg RRA -n 1 -i TEK14.txt -w 128 -p 4 -a 4 -g Sequitur --strategy none
\end{lstlisting}
\end{table}
Following this choice, the advantage of RRA in respect to HOT SAX becomes more limited. 
According to our numerical experiments (Tab. \ref{tab:rraVhs} and Tab. \ref{tab:hsVhs}) in some cases, RRA is even slower than our version of HOT SAX. Notice however that it is always faster than the version of HOT SAX released by one of the authors of RRA, available in the same suite Grammarviz 3.0 \citep{gram2site} and not reported here. 
We utilize the last version of RRA present in the Grammarviz 3.0 suite on GitHub dating the 27th of October 2016.
For completeness, Listing 3 contains the script used to produce RRA results for the time series TEK 14.
When possible we keep the same SAX parametrization  used for the validation of RRA.  They correspond to optimal solutions where the lengths of the sequences are close to the values returned automatically by RRA.
Our code, however, requires that the number of parts of the PAA is an exact divisor of the length of the sequences. If the parameters used for the validation of RRA do not comply with this condition we use values of $P$ which are as close as possible.

The results of Tab. \ref{tab:rraVhs} show that HST outperforms RRA. In the case of ECG 300, HST is about $30$ times faster than RRA, while in the worst-case scenario, for the Daily Communications time series it is only 49\% faster. These improvements are calculated as the ratio of the number of distance calls of RRA and HST, and denoted as D-speedup.  Calculating the ratio of the run times for the two algorithms (T-speedup), does not provide useful information since RRA has been developed in a substantially slower programming language: JAVA (while HST has been developed in Fortran).

\subsection{Disk-Aware Discord Discovery}
\begin{table}[!ht]
\begin{centering}
\begin{tabular}{c|cc|c||cc|c|}

               & \multicolumn{3}{c||}{\small{0.99 r}}   &   \multicolumn{3}{c|}{\small{ exact r}}   \\

               &   \multicolumn{2}{c|}{\small{Runtimes [s]}}  &              &   \multicolumn{2}{c|}{\small{Runtimes [s]}}                           &                 \\ 
 dataset       &     DADD  &     HST    &  T-speedup   &    DADD        &     HST    &  T-speedup      \\
  \hline
Daily commute  &   10.29   &    0.69   &  14.91       &    10.20      &    0.69     &      14.80      \\
Dutch Power    &    7.42   &    0.59   &  12.60       &    7.02      &    0.59     &      11.92      \\
ECG  15        &   17.10   &    0.72   &  23.84       &    9.63      &    0.72     &      13.43      \\
ECG 108        &   11.81   &    0.61   &  19.51       &     8.76     &    0.61     &      14.47      \\
ECG 300        &    8.05   &    0.43   &  18.76       &     6.72     &    0.43     &      15.66      \\
ECG 318        &    6.65   &    0.47   &  14.20       &     6.22     &    0.47     &      13.29      \\
NPRS 44        &   10.82   &    0.55   &  19.71       &   10.71      &    0.55     &      19.50      \\
Video          &   15.25   &    0.60   &  25.37       &   14.91      &    0.60     &      24.80      \\
\end{tabular}
\caption{The second and third columns show the calculation times (seconds) for 10 discords for DADD and HST. The fourth column displays the time-speedup calculated as the ratio of the DADD and HST running times.
The page to be analysed has been built by taking the first 10000 sequences of length 512 of each dataset of column 1.
DADD calculations depend on the $r$ parameter, the timings of column 2 refer to an $r$ value equivalent to 99\% of the $nnd$ of the 10th discord, while the timings of column 5 are associated with the exact value (100\%). 
} \label{tab:dadd}
\end{centering}
\end{table}

DADD (or DRAG)  is known to be a very fast algorithm for discord search aimed at those cases where the whole time series cannot reside on the RAM and requires to be stored on a disk.
For completeness, we report here some running time comparisons with HST (although the nature of the two algorithms is rather different).
The version of DADD used for these tests is a freely available C++ code by \cite{DADD}.
This code is expected to process non-overlapping sequences arranged in pages one after another. For this reason, the concept of self match becomes irrelevant and it has not been implemented. The algorithm computes the Euclidean distance between sequences, without Z-normalization (but the sequences can be z-normalized before being processed by DADD). Each page contains $10^4$ sequences of length 512 points. Given the different data formats processed by HST and DADD, we had to re-arrange the time series we used for the comparison in order to obtain coinciding results.
From the datasets of Tab. \ref{tab:rraVhs} we selected those with more than $10511$ points, and we created one page of 10000 sequences for each of them (the  datasets that did not contain enough points to create one page and were discarded). 
The pages thus formed contain overlapping sequences, however the aim of these experiments is to pinpoint the execution speed only and to obtain coinciding results.

DADD is a two-step algorithm, the first step is aimed at building a restricted pool of sequences with an $nnd$ higher than a certain threshold $r$ (discord defining range). During the second phase, the discords are searched among the sequences selected in the first phase. The value $r$ needs to be imputed at the beginning of the calculation and it affects the calculation time. For example, if one selects a low $r$ value, the first phase will return many sequences, slowing down the second phase. The opposite can also happen, selecting a high $r$ value might lead to a restricted pool that does not contain all the required discords (those with a $nnd>r$). In such a scenario these discords cannot be found and require another calculation with a smaller $r$ value.  
 The value of $r$ is usually obtained by sampling a subset of all of the sequences (for example 1/100 of the sequences). The first k-discords are then obtained from the sample (for the calculations of this section case we choose $k=10$) with the help of a fast discord algorithm. The nearest neighbor distance for the 10th discord from the sample is used as the $r$ parameter for DADD. One does not expect to obtain the exact $nnd$ of the k-th discord but just an approximation. 
Unfortunately, this sampling procedure does not work particularly well for the datasets under consideration. The returned r-parameters are usually too big and require manual adjustments. As a solution, we performed a full calculation on the whole page and we obtained the exact $nnd$ value of the 10th discord. We performed two kinds of calculations, in one case using the exact value and in the second 99\% of it. 
%%%%%%5 new %%%%%%%%
The reason is that the r-parameter affects the computational times for DADD, in general the more distant it is from the exact $nnd$ of the k-th discord, the slower the code. In particular, even a small modification from the exact $r$ to $0.99 r$ can have a non-negligible impact on the calculation times. 
For example, on one dataset (ECG 15) a $1\%$ difference of $r$ leads to a $77\%$ increase of the runtime.
%%%%%%% end new %%%%%%
Although  obtaining this quantity requires some time we do not include  it in the total computational time for the experiments we performed, focusing on the algorithm only.

We first verified the correspondence of the results (we turned off z-normalization in the HST code and we forced the distance calls between overlapping sequences to be allowed). With these modifications, the two codes produced identical results and we could compare the runtimes.
In all the tests, HST is 12-25 times faster than DADD (see Tab. \ref{tab:dadd}). %T

\subsection{SCAMP}\label{ssec:scamp}
\begin{figure}[!ht]
 \includegraphics[width=0.49\textwidth]{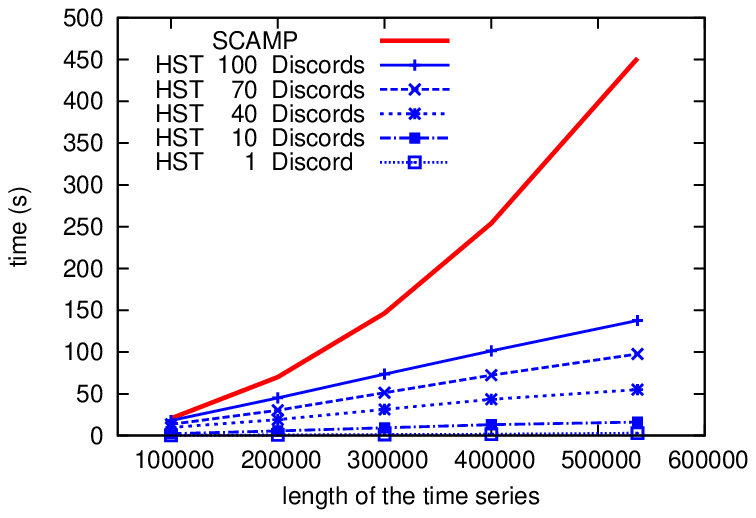}
 \includegraphics[width=0.49\textwidth]{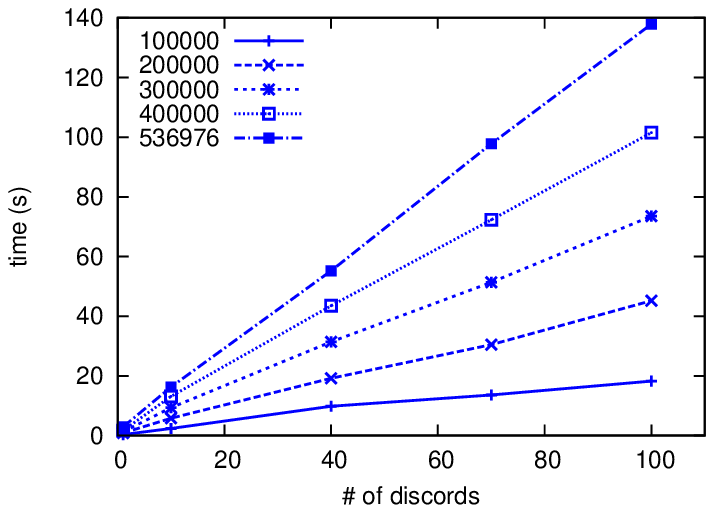}
 \caption{(Left) Calculations performed on sections of the ECG 300 dataset. HST has been run using the same SAX parameters presented in Tab. \ref{tab:rraVhs}l; for each slice of the time series five calculations were performed, for: 1, 10, 40, 70, and 100 discords. The runtimes on the same datasets for SCAMP include only the calculation of the matrix profile. 
 (Right)
 HST runtimes as a function of the number of discords searched for different slices of the ECG 300 time series.
 } \label{fig:HSTvsSCAMP}
\end{figure}
SCAMP is the fastest algorithm of the Matrix Profile series. This algorithm has a broad purpose and it not specialized for discord search, however one of its peculiarities is that it is insensitive in respect to the length of the sequences. Another interesting feature is that looking for many discords is computationally inexpensive once the matrix profile is known. It is difficult to compare an exact algorithm (SCAMP) with a heuristic one (HST), however we provide here some running conditions which can help in deciding when to use one or the other.  
For the tests of this section, we use a C++ code released by the authors of the algorithm, available on GitHub \citep{sitoUCR}. This implementation is capable of exploiting multiple cores and graphic cards, however for a fair comparison with HST we report here the values obtained for a single core (in this case it is essentially identical to another matrix profile algorithm: STOMP \citep{STOMP}). 
SCAMP running times scale quadratically with the length of the time series, but do not depend on the underlying data. At variance, the running times for HST are affected by the characteristics of the specific dataset.
For an indicative comparison we consider parts of the time series ECG 300.
The calculations of Fig.~\ref{fig:HSTvsSCAMP} refer to the dataset truncated at different lengths (we used the same parameters of Tab. \ref{tab:hsVhs}).
The slices contain $10^5$, $2\cdot 10^5$, $3\cdot 10^{5}$, $4\cdot 10^{5}$, and 536976 points (the full time series); for each of these slices we check the running times for HST for 1, 10, 40, 70 and 100 discords. The timings for SCAMP refer only to the calculation of the matrix profile we do not include the time spent for the search of the discords.
For all of the above tests, HST is (much) faster than SCAMP (Fig. \ref{fig:HSTvsSCAMP}) . Moreover since HST scales linearly with the size of the time series, its advantage over SCAMP grows for the longer runs.

For very short time series, SCAMP can become faster than HST. We performed experiments with ECG 0606 (2299 points). In this case, the advantage of SCAMP over HST is so small that the total execution time is dominated by writing the results on the disc. For a slightly longer time series (TEK 14, 5000 points) the two algorithms show comparable times.
We also notice that short time series can have only a few discords, because of the non-overlap condition.

SCAMP can become faster than HST if one is interested in finding hundreds of (long) discords. However, only longer time series can contain so many discords, so the advantage due to the insensitivity to the number of discords is hindered by the quadratic nature of SCAMP. Moreover, one should take into account that only a few of the discords are expected to be ``real'' anomalies.
Some time series do not contain anomalies, while all of them contain many discords: $O(N/s)$.
%
%, 
%%%%%%%%%%%%%% new %%%%%%%%%%%
The reason is simple: discords are just maxima of the matrix profile. Some of these maxima might be abnormally higher than the rest of the matrix profile, thus being anomalies. Others might be just random fluctuations. One is most likely interested only in those discords which are also outliers:  \textit{significant discord}s \citep{sod}. For example, ECG 300 has only 5 \textit{significant discords} of length $s=300$. In this case, it would be useless to calculate the position of the first 100 discords.
%%%%%%%%%%%%%% end new %%%%%%%

%
%

It should be noted that there is an approximate algorithm of the Matrix Profile series called preSCRIMP \citep{scrimp} which produces an approximate version of the matrix profile and which can be used for approximate discords search. Since its results depend on the quality of the approximation and it displays a quadratic complexity with the number of sequences and it is not a direct competitor of HST which is exact.

\subsection{More than a hundred million points time series}\label{ssec:170M}
As an example, we performed a calculation on a time series consisting of 
170~326~411 points associated with the research on insect vector feeding by \cite{insect}. The total computational time for the first 10 discords on an Intel(R) Xeon(R) CPU E5-2640 v3 @ 2.60GHz has been: 
\verb|96288.93| s
(less than 1 day and 3 hours). The parameters in use were $s=512$, $P=128$, $alphabet = 4$. 

This timing is comparable with the fastest exact Matrix Profile parallel implementations exploiting graphic cards \citep{scrimp, scamp}, while the serial implementations would be hundreds of times slower. 
One can also compare the performance of HST and HOT SAX. In this case, we limit the calculations to $k=1$. We obtain a D-speedup = 21, while the T-speedup is 16. For HST the $cps$ is 79, while for HOT SAX it is 1547.

\subsection{HST Scaling}\label{sec:scaling}
\begin{figure}[th!]
 \includegraphics[width=0.499\textwidth]{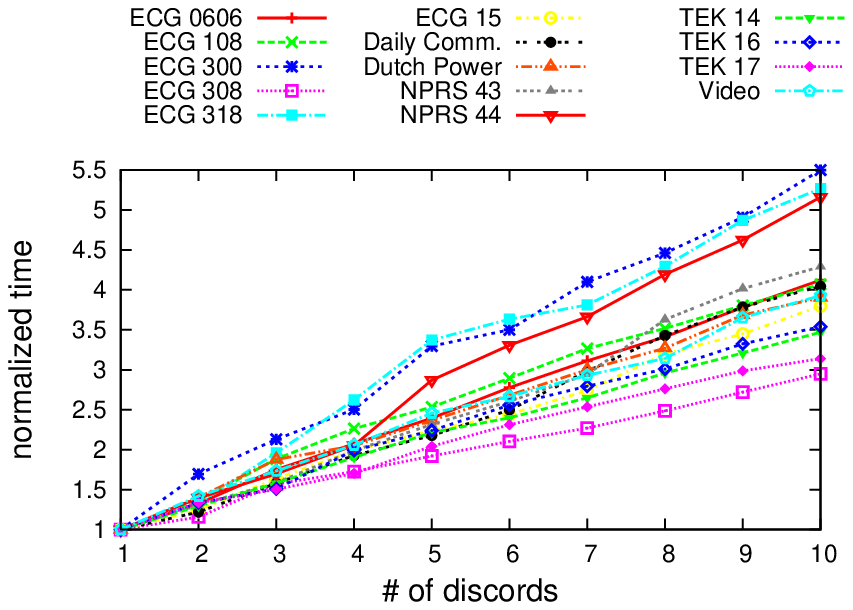}
 \includegraphics[width=0.499\textwidth]{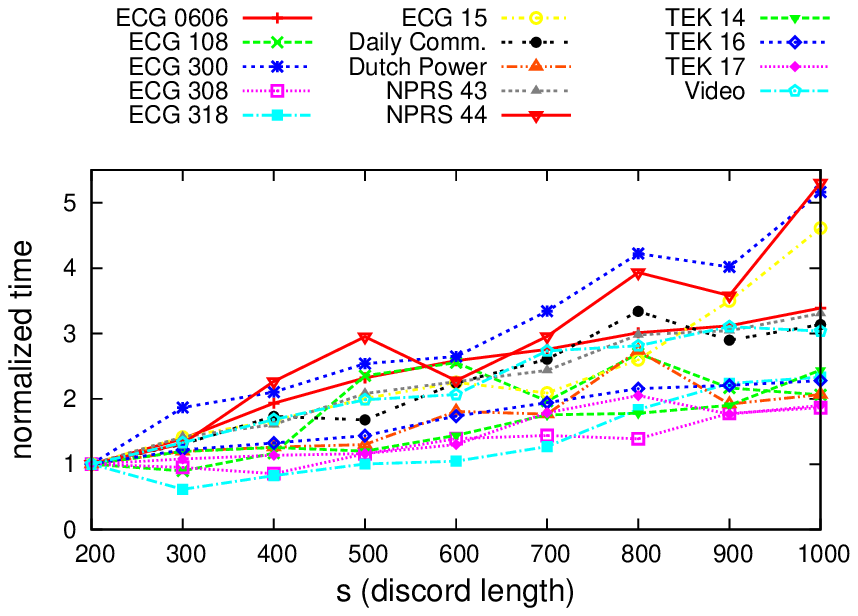}
 \caption{(Left) Running times as a function of the number of discords searched, normalized with the time of the first discord. The length of the sequences in these cases is $s=100$. (Right) The running times for the search of the first discord as a function of the length of the sequences $s$. These values have been normalized with the running times of each dataset for $s=200$.}\label{fig:normalizedTimes}
\end{figure}

In this section, we provide indications regarding how HST scales as a function of the parameters, in many ``normal'' conditions.
The datasets of Tab. \ref{tab:rraVhs} include biological data (the ECG cardiograms and NPRS respiration time series), data regarding human activities (Daily commute, Dutch power and Video), and sensor data (the Shuttle TEK series). They also span a couple of orders of magnitude in terms of length (ranging from $2299$ to $586086$ points). This diversity can be used to gain insights regarding the costs of the calculations as a function of their parameters on ``typical'' time series.
In the beginning, we checked the speed as a function of the quantity of discords to be found. 
%F
Since the running times span a couple of orders of magnitude we normalized the results of each dataset with the running time for the first discord. In Fig. \ref{fig:normalizedTimes} (left) it is possible to notice that in all the cases the curves associated with each dataset are almost linear. 

The dependence of the running times as a function of the length of the sequences is displayed in Fig. \ref{fig:normalizedTimes} (right), for this graph we normalize the values according to the calculations for discords of length $200$. Also in this case the scaling is limited and essentially proportional to the length of the sequences $s$.

The last factor determining the runtimes is the length of the time series. The experiments on ECG 300 show (Fig. \ref{fig:HSTvsSCAMP}, left) that HST runtimes are approximately proportional also to this quantity. 
In summary, HST scales approximately linearly with:
\begin{itemize}
\item the number of discords searched ($k$)
\item the length of the sequences ($s$)
\item the length of the time series ($N$)
\end{itemize} 
These scaling properties suggest a rule of thumb that can be used when facing a very long time series. It is enough to first perform a check on a short extract (e.g. $10^6$ points) and then extrapolate the total running times multiplying the \textit{cps}, the length of the time series and the number of discords searched, to obtain a rough estimate of the total number of calls. Although this recipe is very dependent on the premise that the mechanism producing the time series does not change significantly with time (a non-trivial condition), the results of Fig. \ref{fig:HSTvsSCAMP} (left) seem to indicate that it can be useful.

\subsection{Summary of the results}\label{ssec:analResult}
The tests that we performed show that HST can be an important resource at the time of finding discords when the task is long and complex: 
\begin{itemize}
 \item it is  2-100+ times faster than HOT SAX, it is particularly faster when looking for longer discords or for ``simple looking'' time series. 
 \item it is  50\%-30 times faster than RRA (and HST returns exact discords)
 \item it is 12-25 times faster than DADD (DRAG) 
 \item HST, in practical cases, is much faster than SCAMP and it can produce results comparable to GPU accelerated Matrix Profile algorithms on longer time series.
\end{itemize}

\section{Conclusions and future works}\label{sec:conclusions}
In the present paper, we detail a new algorithm for exact discord search in time series, called HOT SAX Time.  HST can be obtained from HOT SAX with rather easy modifications. 
In HST the external and inner loops are re-arranged following good quality approximate $nnd$s.
During the execution of the algorithm, the order of the sequences of the external loop is rearranged every time that a good discord candidate is found: sequences with a high approximate $nnd$s are positioned at the beginning.

The performance of HST has been validated with real-life and synthetic time series and compared with the results of HOT SAX, RRA, DADD.  HST has shown to be faster than these algorithms for all of the datasets under consideration.
The only serial competitor for HST is SCAMP (running on a single core) in very special cases. The quadratic nature of the matrix profile codes hinders their performance on long time series. For more common cases, HST is many times faster than SCAMP. For long time series ($>$100  million points) the present serial HST code is essentially as fast as the parallel implementation of SCAMP exploiting graphic cards.

Since HOT SAX can be considered as a benchmark algorithm we used it to understand under which condition a discord search becomes complex.
For this purpose, we defined a complexity indicator, the cost per sequence, which allows one to compare searches on time series of different lengths.
We singled out two main parameters that render discord searches particularly complex for HOT SAX. 
The first one is a counter-intuitive property: low noise/signal ratios.
The second parameter with a strong influence on the cost per sequence is the length of the discords. For searches involving long sequences, HST tends to be much faster than HOT SAX; in light of the fact that the technological improvements lead to sensors capable of producing higher sampling frequencies, the advantage provided by HST should grow with time. In particular, to the best knowledge of the authors, HST is the first exact discord algorithm that can be more than 100 times faster than HOT SAX. 

Future developments of this research include more extensive tests in order to better define the characteristics of time series.
It could also be interesting to use alternatives to SAX in order to improve the warm-up phase.
 Parallelizing HST is also a natural follow up of the present work.

\section*{Acknowledgments}
PA would like to thank Prof. M. Zymbler for suggestions regarding DADD, and Prof. E. Keogh for providing the dataset used for Sec. \ref{ssec:170M} and for the useful information regarding the Matrix Profile.

\bibliographystyle{spbasic}      % basic style, author-year citations
\bibliography{biblio}

% \bibliographystyle{apalike}
% {\small
% \bibliography{biblio}}

\end{document}